\pdfoutput=1

\documentclass[11pt]{article}

\usepackage[final]{acl}
\usepackage{times}
\usepackage{latexsym}

\usepackage[T1]{fontenc}

\usepackage[utf8]{inputenc}
\usepackage{xcolor}

\usepackage{microtype}

\usepackage{inconsolata}
\usepackage{times}
\usepackage{latexsym}
\usepackage{tabularx}
\usepackage{multirow}
\usepackage{booktabs}
\usepackage{makecell}
\usepackage{graphicx}
\usepackage{hyperref} 
\usepackage{amsmath} 
\usepackage{float}
\usepackage{soul}
\usepackage[utf8]{inputenc}
\usepackage{titlesec}
\usepackage[T1]{fontenc}

\usepackage[utf8]{inputenc}

\usepackage{microtype}
\usepackage{algorithmic}
\usepackage[linesnumbered,ruled,vlined]{algorithm2e}
\usepackage{inconsolata}

\usepackage{cleveref}
\usepackage{textgreek}
\usepackage{tipa}
\usepackage{multirow} 
\usepackage{xcolor}
\usepackage{soul}  

\crefname{section}{§}{§§}
\usepackage{graphicx}
\usepackage{tikz}
\usepackage[most]{tcolorbox}
\usepackage{titlesec}

\titlespacing*{\subsection}
{0pt}           
{12pt plus 4pt minus 2pt}  
{6pt plus 2pt minus 2pt} 
\newtcolorbox[list inside=prompt,auto counter,number within=section]{prompt}[1][]{
    colbacktitle=black!60,
    fonttitle=\small,
    coltitle=white,
    fontupper=\footnotesize,
    boxsep=4pt,
    left=0pt,
    right=0pt,
    top=0pt,
    bottom=0pt,
    boxrule=1pt,
    #1,
}
\title{Take Out Your Calculators: Estimating the Real Difficulty of \\Question Items with LLM Student Simulations}

\author{Christabel Acquaye,~~~Yi Ting Huang,~~~Marine Carpuat,~~~Rachel Rudinger \\
    University of Maryland, College Park \\
    \texttt{\{cacquaye, ythuang1, marine, rudinger\}@umd.edu} \\
}

\begin{document}
\maketitle
\begin{abstract}
Standardized math assessments require expensive human pilot studies to establish the difficulty of test items. We investigate the predictive value of open-source large language models (LLMs) for evaluating the difficulty of multiple-choice math questions for real-world students. We show that, while LLMs are poor direct judges of problem difficulty, simulation-based approaches with LLMs yield promising results under the right conditions. Under the proposed approach, we simulate a ``classroom'' of 4th, 8th, or 12th grade students by prompting the LLM to role-play students of varying proficiency levels. We use the outcomes of these simulations to fit Item Response Theory (IRT) models, comparing learned difficulty parameters for items to their real-world difficulties, as determined by item-level statistics furnished by the National Assessment of Educational Progress (NAEP). We observe correlations as high as 0.75, 0.76, and 0.82 for grades 4, 8, and 12, respectively on the item-level correctness rates. In our simulations, we experiment on math MCQs with different ``classroom sizes,'' showing tradeoffs between computation size and accuracy. We find that role-plays with diverse named students improves predictions (compared to student ids), and stratifying names across gender and race further improves predictions. Our results show that LLMs with relatively weaker mathematical abilities (Gemma) actually yield better real-world difficulty predictions than mathematically stronger models (Llama and Qwen), further underscoring the suitability of these models for the task.





\end{abstract}

\begin{figure}[t]
    \centering
    \includegraphics[width=0.85\linewidth]{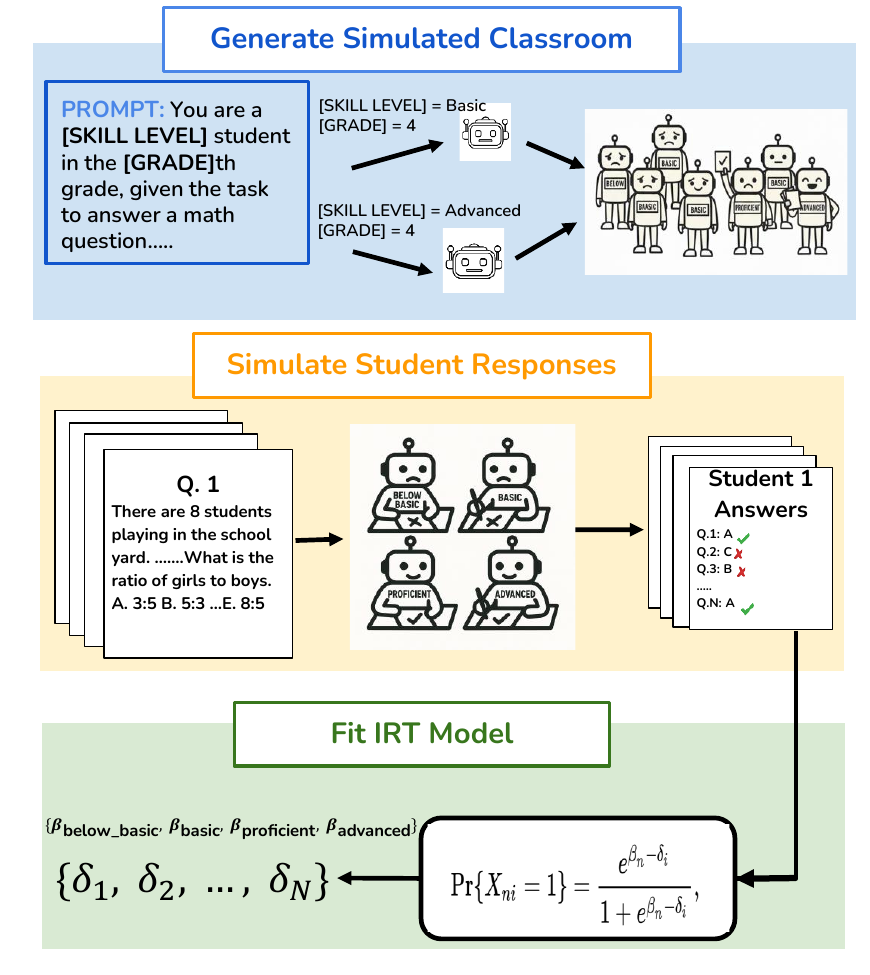}
    \caption{We simulate classrooms by prompting LLMs at different skill levels. Simulated students with varying abilities respond to question items . An IRT model is fit to the simulated responses to estimate students' ability, $\beta$ and item difficulty, $\delta$.}
    \label{fig:pipeline}
\end{figure}

\section{Introduction}
The accurate estimation of question difficulty is key for developing reliable educational assessments as difficulty is essential in determining the quality and validity of these test items \citep{Alkhuzaey2023TextbasedQD}. Multiple-choice mathematics question items which measure mathematical ability independent of contextual knowledge, requires rigorous validation of difficulty to be used in assessments. Traditional approaches rely on expert cognitive analysis \citep{Leight, CDM} or retrospective psychometric studies of student performance \citep{Harris, bond2013applying} through pilots. However, these pilot studies are costly, time-consuming, and difficult to scale with costs reaching thousands of dollars and weeks to complete \citep{NationalAcademies2022}.

Automated approaches like Question Difficulty Estimation from Text (QDET) offer a potential solution, but produce weak correlations with real world difficulty estimation\citep{scarlatos2025smartsimulatedstudentsaligned}. Even with chain-of-thought and stronger closed-source models, direct text-based prediction fails to capture the misconception patterns underlying item difficulty \citep{li-etal-2025-item}. As difficulty inherently depends on who answers, not just what is asked \cite{MartnezPlumed2022WhenAD}, we need individual-level responses for established framework Item Response Theory \citep{Hambleton1985IRT} to estimate both difficulty and discriminability parameters, that direct estimation cannot provide. This requires costly individual response data \citep{Zelikman2023GeneratingAE}. Even when such data exist, they are often proprietary or provide only aggregate metrics rather than the individual response IRT requires, thus a need to simulate realistic student responses.

Recent work shows that LLMs can simulate human-like responses across diverse tasks, including preliminary explorations in assessments \citep{lu2024generative, Xu_2025, liu2023gptbasedopenendedknowledgetracing}. We extend this direction by prompting open-source models, requiring no additional training data/fine-tuning, and validate difficulty estimates using IRT.

We simulate diverse student profiles (varying skill levels from Below Basic, Basic, Proficient, Advanced and identity attributes including student ids and demographic names) and validate these simulations against real NAEP mathematics data using IRT (Figure~\ref{fig:pipeline}). We study:
\textbf{(1)} To what extent do LLM-simulated student responses mimic real-world difficulty \textbf{(2)} How do prompt design, identity attributes and model selection affect simulated vs real student performance alignment? We find correlations as high as 0.75, 0.76, and 0.82 for grades 4, 8, and 12, respectively, with LLMs with relatively weaker mathematical abilities and an ensemble of top performing models falling within an acceptable range by standard reliability benchmarks  \citep{Cicchetti1994Guidelines}. Diverse identity improves correlations over no-identifier baselines and IRT-estimated difficulties achieve AUC scores of 0.77--0.90, suggesting that simulations can discriminate between items that real students find easy versus hard. LLMs with relatively weaker mathematical abilities outperform highly-proficient models, suggesting that these proficient models struggle to reproduce realistic error patterns. These findings show open source LLM simulations can offer a practical low-cost screening tool, with the counterintuitive finding that LLMs with relatively weaker mathematical abilities often better simulate struggling students.
\section{Related Work}
Recent work show LLMs can act as ``silicon'' subjects, reproducing human behavioral patterns across diverse tasks \citep{xie2024can, Argyle_Busby_Fulda_Gubler_Rytting_Wingate_2023, DILLION2023597, manning2024automated, yang2024social}. \citet{scarlatos2025smartsimulatedstudentsaligned} propose SMART for difficulty prediction of open ended question using Direct Preference Optimization (DPO) and IRT, while \citet{lu2024generative} introduce a \textit{Generative Students} framework where GPT-4 is prompted with student knowledge profiles to answer Multiple Choice Questions (MCQs), finding that the LLM responses align with the intended profiles and that the set of “hard” questions align with real students' difficulty. Similarly, \citet{benedetto-etal-2024-using} develop skill-level prompts for GPT-3.5/4 across domains, noting model-specific tuning requirements. \citet{liu2024leveragingllmrespondentsitemevaluation}  use multiple LLMs (GPT-3.5/4, Llama 2/3, Gemini-Pro, etc.) to pretest College Algebra items.  \citet{Feng} use a supervised approach for difficulty prediction by training models using ground-truth IRT difficulty estimates derived from real student responses. Other works consider incorporating student learning behaviors, knowledge states, and memory limitations into LLM simulations, to provide potential alternatives to conventional knowledge tracing systems \citep{Wang2023ASO,Hu2025ExploringTP}. Our work extends these with open source models, by estimating abilities at a group level where we do not have individual real world student responses, examining student ids and first names as demographic identifiers in our prompting techniques and leveraging ensembling techniques across models.

\section{Preliminaries}
\subsection{Data}
We collect 631 math MCQs from the National Report Card website \cite{nces2021math} for grades 4, 8 and 12. NAEP is a congress authorized project of the National Center for Education Statistics (NCES) with the Institute of Education Sciences of the U.S. Department of Education. These questions provide real student statistics across the United States, providing a benchmark against which our LLM simulations can be evaluated. This data also serves as the only nationally representative and continuing assessment of student achievement in the USA. Questions from the NAEP have undergone rigorous development and validation processes. Each problem includes a multiple choice question, several answer choices, a correct answer, and metadata including difficulty, content area, grade level, and student performance statistics across demographics. The items cover a diverse range of mathematical concepts. Figure~\ref{fig:question_stats} in the appendix presents the distribution of question items across content areas.
Given that we are not conducting combinatorial experiments across multiple variables, but rather evaluating LLM capabilities against human benchmarks, this sample size provides sufficient coverage of key mathematical concepts. Table~\ref{tab:difficulty_gender_breakdown} in the appendix summarizes the distribution of question items by grade level and difficulty. \footnote{Data is available at \url{https://github.com/christabel-acquaye/simulated_students}}

\subsection{Task}
Our task is to use LLMs to predict classroom performance by simulating how $N$ students with varying skill levels would answer a given math question. Each simulated response is graded as correct or incorrect against the answer key as shown in Figure~\ref{fig:pipeline}. Given a math question, $q$ with multiple answer choices $\{a_1, a_2..a_m\}$, we generate $N$ individual student responses. We then aggregate these responses to compute the simulated classroom success rate as an estimated proportion $\hat{y}_q \in [0,1]$ representing the predicted percentage of simulated students who would correctly answer the question $q$. To evaluate our approach, we test whether simulated performance aligns with real-world data. Given a set of $Q = \{q_1, q_2, ..., q_m\}$ with student performance statistics $Y = \{y_1, y_2, ..., y_m\}$ (where each $y_i \in [0,1]$ represents the actual percentage of students answering $q_i$ correctly), we compute the correlation $r(\hat{Y}, Y)$, between our simulated success rates $\hat{Y}$ and real world success rates, $Y$. In addition to percentage correct rates, we ground our approach using Item Response Theory (IRT), a well established psychometric framework that estimates item difficulties ($\delta$) and student abilities ($\beta$) by modeling response probabilities. IRT requires individual student responses, to separate true item characteristics from population-level effects. We fit our IRT model to our simulated $N$ students responses and test whether the estimated difficulty parameters predict real world outcomes, showing psychometric validity beyond simple correlations.

\section{Experiment }

We extend \citet{benedetto-etal-2024-using} and \citet{lu2024generative}, to generate simulated students with varying skill competencies. We map each simulated student to a grade level, and one of the four National Assessment of Educational Progress (NAEP) levels: Below Basic, Basic, Proficient, or Advanced.
 These NAEP levels provide the concrete descriptors for the skills and performance we attribute to each simulated student.\footnote{The NAEP definitions for the performance levels are described \href{https://nces.ed.gov/nationsreportcard/mathematics/achieve.aspx}{here}.}
We also simulate students using student ids (\ref{sec:UIDs}) and first names to test whether individuating students further improves simulation quality. To avoid overgeneralization to a specific demographic, we take a stratified sample of names across intersections of associated gender and race/ethnicity. For each question, we generate responses for a simulated classroom of size $N$ with a non-uniform skill distribution: 25\% Below Basic, 35\% Basic, 25\% Proficient, and 15\% Advanced, reflecting NAEP’s typical pattern of a large Basic cohort, roughly equal Below-Basic and Proficient groups, and a small Advanced group.

\textbf{Names} To investigate whether adding identity markers improves simulation, we assign first names to simulated students. Prior work shows name embeddings are sensitive to demographic correlates of first names \cite{caliskan2017semantics, wolfe-caliskan-2021-low,zhang2024climb, acquaye-etal-2024-susu}.
We sample 48 names evenly distributed between four associated races (Asian, Black, Hispanic, and White) and genders (female and male) to reduce chances of over-generalization of the simulation on a single name, demographic association, or stereotype. While \textit{individual} simulations may exhibit demographic-based stereotypes, in this work our goal is to use \textit{aggregate} statistics to predict entire \textit{population}-level outcomes on test items, though future work can look at distinguishing this stereotype activation. We emphasize that we do not use these names to make predictions about specific people/groups but report empirical observations of using these names. These names were selected following \citet{an-etal-2024-large}.  A comprehensive list of these names is in~\cref{sec:appendix_first_names}. 

\begin{table}[h!]
\centering
\resizebox{\columnwidth}{!}{%
\begin{tabular}{lccccccccccc}
\toprule
 & \multicolumn{2}{c}{\textbf{G2}} & \multicolumn{3}{c}{\textbf{G3}} & \multicolumn{2}{c}{\textbf{L3}} & \multicolumn{2}{c}{\textbf{Q2.5}} & \textbf{Q3} \\
\cmidrule(lr){2-3} \cmidrule(lr){4-6} \cmidrule(lr){7-8} \cmidrule(lr){9-10} \cmidrule(lr){11-11}
 & \textbf{9B} & \textbf{27B} & \textbf{4B} & \textbf{12B} & \textbf{27B} & \textbf{8B} & \textbf{70B} & \textbf{14B} & \textbf{32B} & \textbf{4B} \\
\midrule
\textbf{Acc. $\uparrow$} & 0.72 & 0.74 & 0.56 & 0.78 & 0.83 & 0.81 & 0.92 & 0.92 & \textbf{0.93} & \textbf{0.93} \\
\bottomrule
\end{tabular}%
}
\caption{Model Accuracy Across All Questions. G2/G3 = Gemma-2/3, L3 = Llama-3, Q2.5/Q3 = Qwen-2.5/3.}
\label{tab:overall_accuracy}
\end{table}
\textbf{Models} We experiment with ten open-source LLM from different model families and generations of varying sizes: Gemma-2 (9b and 27b) \cite{gemma_2024}, Gemma-3 (4b, 12b, 27b) \cite{gemma_2025}, Llama 3.1-8B and Llama-3.3-70B \citep{llama3}, Qwen2.5 (14B and 32B) \cite{qwen25} and Qwen3-4B\cite{qwen3technicalreport}. These models show moderate to good math reasoning with accuracy scores (Table \ref{tab:overall_accuracy}) on the NAEP questions, as well as abilities in role-playing and following instructions. Recent works study open source LLMs in replicating human behavior, cognition and testing settings \cite{schröder2025largelanguagemodelssimulate, scarlatos2025smartsimulatedstudentsaligned, Xu2025HumanlikeCR, Jiang2025RepresentationCF} which makes them candidates for our student simulations. We \textbf{note} that all selected models are instruction-tuned and used according to their intended purpose. However, we recognize that a model's ability to simulate different student performance (such as an advanced student who answers more correctly versus a below-basic student) will be inherently influenced by whether the model can actually solve the question correctly on its own, as shown in Table \ref{tab:overall_accuracy}.

\subsection{Methods}

\textbf{Direct Question Difficulty Estimation from Text} Recent Question Difficulty Estimation from Text (QDET) \cite{Benedetto2023AQS} approaches use embeddings or transformers to directly estimate the difficulty of question items. We implemented Word2Vec-based difficulty estimation, which encodes question texts as embeddings and uses them as input to a Random Forest regression model to output continuous difficulty values. For transformer baselines, we extract [CLS] token embeddings from pre-trained BERT-base-uncased and DistilBERT-base-uncased and run inference on question text concatenated with the correct answer. This setting enables comparison with LLM prompting, where correlations are computed over all 631 items without task-specific training.

\noindent\textbf{Direct Percentage Correct Estimation (DPCE)}
We also establish baseline performance by directly prompting LLMs to estimate the percentage of students, at a specified grade level, who would solve the given question item correctly with prompt~\ref{prompt:direct_prompt}. This tests whether models can directly assess a given question item's difficulty by predicting students success rates (i.e, percentage of students who answer correctly) to estimate how challenging the question item would be for students. The idea is that questions answered correctly by most students are estimated as easier while those answered incorrectly by most students are estimated as harder.  We run this by first setting our (temp. $T = 0$) and using greedy decoding to generate a single deterministic single prediction for the percentage of student who would answer correctly for each question. The second averaged approach, uses stochastic sampling (temp. $T = 0.3$) to generate ten responses per question, then aggregates their predictions by averaging the resulting percentages. We theorize that simulating multiple students responses is a more reliable way to get information about question difficulty from LLMs rather than asking them directly \cite{scarlatos2025smartsimulatedstudentsaligned}, as difficulty inherently depends on who answers, not just what is asked, however the latter is computationally cheaper, thus we include it as a baseline. 

\noindent\textbf{Simulated Classroom Performance Estimation}
    We prompt the LLMs in a student role-play prompt approach to generate $N$ simulated student responses. Without simulating individual students, we cannot access the variance patterns and response distributions that reveal whether a question is genuinely difficult. A single prediction (e.g., "60\% of students will answer correctly") from DPCE cannot provide the response variance patterns necessary for difficulty estimation using established psychometric IRT framework. The prompt asks the LLM to answer the question as a student of a given skill profile with Prompt~\ref{prompt:student_prompt}. 
For each question, we prompt LLMs to generate individual student responses using role-play prompts that specify grade level and NAEP skill level (Below Basic, Basic, Proficient, Advanced). Each simulated response is scored as correct or incorrect against the answer key. We then compute the classroom success rate: the proportion of all simulated students who answered correctly, aggregated across the skill distribution (25\% Below Basic, 35\% Basic, 25\% Proficient, 15\% Advanced).
To investigate whether student identity grounding affects difficulty estimation, we vary the identifier type across four conditions: (1) No Identifier (2) Student IDs, where each student receives a randomly generated alphanumeric identifier (e.g., STU000142); (3) Single Name, where all students in a classroom share the same first name but differ in skill level. For example, one classroom might consist entirely of students named  `Aryan'  with varying skill levels, while another consists of students named `Nichelle'; and (4) Diverse Names, where students have different first names sampled across demographic groups (gender and race). This setup tests whether diverse student identities in simulated classrooms improves correlation with real-world performance using prompts ~\ref{prompt:demographic_prompt} and ~\ref{prompt:ids_prompt}. 
Including student ids and demographic names serves as a minimal persona cue to empirically improve correlation, though the underlying mechanism remains uncertain. This improvement could reflect increased prompt specificity, or other factors requiring further investigation. This effect is important to produce more consistent simulations that leads to more stable IRT parameter estimates, which in turn produce more reliable difficulty predictions that educators can rely on when designing assessments. We investigate practical considerations, by varying the class size to investigate whether additional simulated responses lead to more reliable difficulty estimates. While increased simulation sizes reduces sampling variance, they also incur higher computational costs. By testing multiple classroom sizes, we aim to understand how correlation strength scales with our class size.


\noindent\textbf{IRT Difficulty Estimation}
We estimate item difficulties using an IRT model \cite{rasch1980probabilistic} fitted to the individual binary response data simulated from each model for our simulated classroom, in order to understand how accurately these estimated item difficulties accurately predict real world outcomes, using the equation below.\[
P\big(X_{ni} = 1 \mid \beta_n, \delta_i\big)
= \frac{\exp(\beta_n - \delta_i)}{1 + \exp(\beta_n - \delta_i)}
\]
where: $\beta_n$ = Ability of student $n$; 
$\delta_i$ = Difficulty of item $i$. 
In our MCQ setting, we compare the simulated students response with the correct answer key and treat each simulated student's response as binary (correct=1, incorrect=0) and fit the IRT model using maximum likelihood estimation to jointly estimate student abilities at a group level ($\{\beta_{\text{below\_basic}}, \beta_{\text{basic}}, \beta_{\text{proficient}}, \beta_{\text{advanced}}\}$) and item difficulties from the response.  After fitting the IRT model to responses from our simulated classroom of diverse students, we extract the estimated difficulties $\delta_i$ for each test item, that we then compare with real world difficulties. When IRT estimated difficulties align with real-world outcomes, it demonstrates that we have captured the true psychometric properties of each item so that a difficult question item remains difficult whether tested on high-performing or struggling students.\footnote{Appendix \ref{sec:appendix_experiments_} summarizes all experiments conducted for this work.}

\subsection{Evaluation}
Before applying IRT to extract difficulty parameters, we must validate that our simulated responses reflect real student behavior, otherwise even the most sophisticated psychometric analysis would yield meaningless results. For each math question, we compute  an accuracy, representing the percentage of students in a simulation that got the question right, which we compare to real-world accuracy rates from NAEP student testing meta-data.

\noindent We compute correlation between simulated and real-world accuracy rates using Pearson (linear relationship) and Spearman (rank ordering). To validate IRT-estimated difficulties, we use regression to predict difficulty class, evaluating with AUC to measure how well difficulties discriminate between easy and hard items. We do this as we do not have direct numerical ground truth difficulty estimates to directly compare our estimated difficulty against. We use Pearson and Spearman correlations rather than error-based metrics such as MSE because we have the aggregate student accuracy rates. This makes correlation with accuracy rates the appropriate primary metric, as it measures the degree to which our simulations preserve the relative ordering of item difficulty.  

\section{Results and Discussion}
We validate the relationship between expert-assigned difficulty labels of (Easy,  Medium and Hard) with real world student outcomes on the NAEP math word problems across different grade levels. Figure~\ref{fig:human_correct} shows \textit{`Easy' } questions consistently give higher average correctness rates for the real world across different grade levels, while \textit{`Hard'} questions show lower correctness rates. This indicates the percentage of students answering a question correctly informs item difficulty. We present baseline results from direct difficulty estimation and simulated students results that show simulating students produces difficulty estimates that correlate better with the real world.
\begin{table*}[h!]
\centering
\resizebox{\textwidth}{!}{%
\begin{tabular}{lccccccccc}
\toprule
\multirow{2}{*}{\textbf{Model}} & \multicolumn{3}{c}{\textbf{(r, PC) $\uparrow$}} & \multicolumn{3}{c}{\textbf{($\rho$, SC) $\uparrow$}} & \multicolumn{3}{c}{\textbf{AUC $\uparrow$}} \\
\cmidrule(lr){2-4} \cmidrule(lr){5-7} \cmidrule(lr){8-10}
 & \textbf{G4} & \textbf{G8} & \textbf{G12} & \textbf{G4} & \textbf{G8} & \textbf{G12} & \textbf{G4} & \textbf{G8} & \textbf{G12}\\
\midrule
Gemma-2-9b  & \textbf{0.69} {\scriptsize [0.61, 0.74]} & 0.61 {\scriptsize [0.52, 0.67]} & 0.74 {\scriptsize [0.63, 0.80]} & \textbf{0.69} {\scriptsize [0.59, 0.74]} & 0.59 {\scriptsize [0.49, 0.66]} & 0.73 {\scriptsize [0.61, 0.81]} & 0.84 & 0.78 & 0.84\\
\midrule
Gemma-3-12b & 0.66 {\scriptsize [0.57, 0.72]} & \textbf{0.70} {\scriptsize [0.64, 0.76]} & 0.76 {\scriptsize [0.67, 0.81]} & \textbf{0.67} {\scriptsize [0.56, 0.74]} & \textbf{0.70} {\scriptsize [0.62, 0.77]} & 0.75 {\scriptsize [0.65, 0.80]} & 0.82 & 0.82 & \textbf{0.88}\\
Gemma-3-27b & \textbf{0.70} {\scriptsize [0.62, 0.76]} & \textbf{0.72} {\scriptsize [0.66, 0.77]} & 0.75 {\scriptsize [0.67, 0.82]} & \textbf{0.74} {\scriptsize [0.66, 0.81]} & \textbf{0.72} {\scriptsize [0.65, 0.78]} & 0.75 {\scriptsize [0.66, 0.83]} & \textbf{0.88} & \textbf{0.85} & \textbf{0.89}\\
\midrule
Llama-3-8b    & 0.42 {\scriptsize [0.32, 0.52]} & 0.44 {\scriptsize [0.34, 0.53]} & 0.44 {\scriptsize [0.31, 0.58]} & 0.40 {\scriptsize [0.28, 0.51]} & 0.46 {\scriptsize [0.36, 0.55]} & 0.48 {\scriptsize [0.32, 0.62]} & 0.67 & 0.72 & 0.75\\
\midrule
Qwen2.5-32B & 0.60 {\scriptsize [0.51, 0.68]} & 0.62 {\scriptsize [0.55, 0.69]} & 0.51 {\scriptsize [0.40, 0.62]} & \textbf{0.64} {\scriptsize [0.54, 0.73]} & \textbf{0.65} {\scriptsize [0.56, 0.72]} & 0.50 {\scriptsize [0.37, 0.61]} & 0.82 & 0.81 & 0.76\\
\midrule
Qwen3-4b & 0.60 {\scriptsize [0.51, 0.68]} & 0.58 {\scriptsize [0.49, 0.66]} & 0.62 {\scriptsize [0.51, 0.72]} & \textbf{0.64} {\scriptsize [0.54, 0.73]} & 0.63 {\scriptsize [0.54, 0.71]} & 0.62 {\scriptsize [0.48, 0.73]} & 0.79 & 0.80 & 0.83\\
\bottomrule
\end{tabular}
}
\caption{Selected Models Correlation and AUC of Simulated Students. All correlations are statistically significant at p < 0.01. Values in parentheses show bootstrap 95\% confidence intervals (n=1000 resamples). \textit{r} = Pearson correlation; $\rho$ = Spearman correlation; AUC = Area Under Curve. See full results including $R^2$ for other models in Table \ref{tab:full_results_all}}
\label{tab:full_results}
\end{table*}
\subsection{Direct Difficulty Estimation Produces Unreliable Predictions}
\noindent \textbf{Question Difficulty Estimation from Text (QDET)} Table~\ref{tab:approach_results} shows Word2Vec has moderate performance 
(ρ=0.36 at Grade 4, 0.30 at Grade 12) but weakest at Grade 8 (ρ=-0.04). 
BERT and DistilBERT have relatively weaker correlations (ρ=0.01-0.21), 
suggesting that these text embeddings provide limited signal 
for our domain-specific difficulty prediction. These results show that question and answer text complexity is insufficient for accurate 
difficulty prediction in our NAEP context.

\noindent \textbf{Direct Prompting LLMs} Correlations between these LLM-generated percentage predictions and real human student percentages shows that direct prompting has weak correlations across different models for different grade levels, as indicated by the negative correlation values reported in Table~\ref{tab:direct_results} and \ref{tab:averaged_results} of the appendix. 
Across all models tested, correlations remain consistently near zero (ranging from -0.139 to 0.137), with no model showing moderate or strong correlations. These predictions show a mix of near-zero positive and negative correlations, with no clear signal from direct prompting at any grade level. .We additionally evaluate GPT-4o under the same direct prompting setup and find GPT-4o produces similarly weak correlations (ρ = 0.01–0.15 across grades), confirming that the failure of direct difficulty estimation is not a function of model capability.

\subsection{LLMs Can Simulate Real Student Correctness Rates}
\noindent \textbf{Simulating LLM students performance correlates with real world students performance.}
Our results provide evidence that LLM-simulated students can mimic response selections relevant for difficulty estimation to an extent. Using a class size of 300 with sampling proportion (75 Below basic, 105 Basic, 75 proficient and 45 advanced students), the aggregated simulated student performance shows some correlation with real-world performance as seen in Table~\ref{tab:full_results}, although the strength varies substantially across models and grades.
Correlation coefficients range from 0.40 to 0.78. We find the Gemma models have the relative strongest correlations, with Gemma-2-27b achieving the highest correlations at grade 12 (r = 0.78, ρ = 0.78), while the Llama models show weaker correlations, with Llama-3-8b ranging from 0.42 to 0.44.

\noindent \textbf{Models that struggle to solve the questions themselves better simulate student performance.} A key finding is that larger and higher predictive accuracy models do not necessarily produce better simulations of student performance. Within the Gemma-2 family, the 9B parameter model achieves comparable correlations to the 27B model at grade 4 (r = 0.69 vs. 0.69), despite having fewer parameters. We also find, Qwen2.5-32B (one of the largest and highest predictive accuracy models tested), produces only moderate correlations (r = 0.51-0.62) suggesting that model scale alone is insufficient for simulation fidelity. Similarly, Llama-3.3-70b shows strong baseline accuracy (92\% accuracy) but produced relatively poor simulation correlations (Table \ref{tab:full_results_all}). This suggests that models that are adept at solving mathematical problems may struggle to authentically simulate the response patterns of struggling students. Models that find the problems more challenging may better capture the distribution of student difficulties. Across grades, we generally observe stronger correlations for higher grades, though this pattern is not universal. Most models show improvement from grade 4 to grade 12, with notable exceptions: Gemma-2-9b exhibits lower correlations at grade 8 (r = 0.61) compared to grades 4 and 12, while Qwen2.5-32B shows worse correlations at grade 12. The Gemma family shows the most consistent performance across grade levels, with Gemma-3-27b maintaining correlations above r = 0.70 and ρ=0.72 (Spearman) across all grades. We also visualize the correlation patterns in scatter plots shown in Figure ~\ref{fig:scatter_plots} which show that Gemma models with stronger correlations have tighter clustering around the diagonal of perfect agreement, with the select Gemma models showing relatively clearer linear trends than Llama.
\begin{table}[h!]
\centering
\small
\resizebox{1\columnwidth}{!}{%
\begin{tabular}{lcccccc}
\toprule
\multirow{2}{*}{\textbf{Approach}} & \multicolumn{3}{c}{\textbf{(r, PC) $\uparrow$}}& \multicolumn{3}{c}{\textbf{($\rho$, SC) $\uparrow$}} \\
\cmidrule(lr){2-4} \cmidrule(lr){5-7}
 & \textbf{G4} & \textbf{G8} & \textbf{G12} & \textbf{G4} & \textbf{G8} & \textbf{G12} \\
\midrule
Averaged & 0.73 & 0.72 & 0.74 & 0.72 & 0.70 & 0.74 \\
Weighted & \textbf{0.75} & \textbf{0.76} & \textbf{0.82} &\textbf{ 0.75} & \textbf{0.75} & \textbf{0.82} \\
\bottomrule
\end{tabular}
}
\caption{Ensemble Model Performance Across Grades}
\label{tab:ensemble_results}
\end{table}

\noindent \textbf{Ensembling models improves correlation strength.}
We ensemble models in Table \ref{tab:ensemble_results} in an average approach (all 10 models percentage correct value is averaged for each item) and in a weighted (equal weights) ensemble of top-performing models (Gemma-2-9b, Gemma-2-27b, Gemma-3-27b).  We see that averaging all model predictions gives ρ=0.70-0.74, while our weighted ensemble of top-performing models  achieves a relatively stronger correlation of ρ=0.82 at Grade 12.

\noindent \textbf{Performance varies by content area.}
Content-area analysis (Table \ref{tab:content_area_results}) show simulations excel at Measurement (ρ=0.75-0.88 across grades for most Gemma models) but struggle with Algebra (ρ=0.45-0.64) at Grade 8 for these Gemma models.

\begin{table}[h!]
\centering
\small
\resizebox{1\columnwidth}{!}{%
\begin{tabular}{lccccccc}
\toprule
\textbf{Class Size} & \multicolumn{3}{c}{\textbf{(r, PC) $\uparrow$}} & \multicolumn{3}{c}{\textbf{($\rho$, SC) $\uparrow$}} \\
\cmidrule(lr){2-4} \cmidrule(lr){5-7}
 & \textbf{G4} & \textbf{G8} & \textbf{G12} & \textbf{G4} & \textbf{G8} & \textbf{G12} \\
\midrule
50 & 0.62 & 0.57 & 0.67 & 0.61 & 0.55 & 0.66 \\
100 & 0.67 & 0.60 & 0.71 & 0.66 & 0.58 & 0.71 \\
300 & \textcolor{black!70!black}{\textbf{0.69}} & \textcolor{black!70!black}{\textbf{0.61}} & \textcolor{black!70!black}{\textbf{0.74}} & \textcolor{black!70!black}{\textbf{0.69}} & \textcolor{black!70!black}{\textbf{0.59}} & \textcolor{black!70!black}{\textbf{0.73}} \\
\bottomrule
\end{tabular}
}
\caption{Correlation values for Gemma-2-9b-it by class size across grades. Larger class sizes consistently improve correlation with real student skill levels. All values statistically significant with p<0.01.}
\label{tab:class_size_results}
\end{table}

\noindent \textbf{Larger class sizes improve reliability.}
We varied the class size for our simulated students while keeping other parameters constant. Using the Gemma 2-9b model for its speed of runs and smaller parameter size, we generate individual student responses for simulated classrooms of 50, 100 and 300 with the same skill level distributions maintained across these sizes. The results in Table~\ref{tab:class_size_results} show a pattern of improved correlations with larger class sizes across all grade levels.
For grade 4, correlations increase from (r = 0.62 to 0.69) as class size grows(50 to 300), while the increase from 100 to 300 students provides only marginal benefits. We note that while larger samples provide more reliable statistical relationships \cite{Luxburg2008StatisticalLT} they require more resources compared to smaller sizes. 
This trade-off indicates a practical workflow where researchers can use small simulations (n=50) for rapid iteration during assessment development, and validate promising items with larger simulations (n=300) before any human testing. 

\noindent \textbf{Diverse name identifiers improves correlation performance.}
We simulate a classroom of 50 students using Gemma-2-9b with different identifier strategies: no identifier, student IDs, single homogeneous names (all students share one name), and diverse names sampled across gender and race. We find that providing student identifiers improves performance over no-identifier baselines, with student IDs (Pearson: 0.60-0.72, Spearman: 0.58-0.72) outperforming no identifier (Pearson: 0.57-0.67, Spearman: 0.55-0.66) across grades(Table~\ref{tab:prompt_name_correlation}). For single homogeneous names (where all 50 students share the same name but have different skill levels), we observe some variance across individual name choices (SD=0.059-0.090), with Grade 12 correlations ranging from 0.48 (Eun) to 0.77 (Inaaya) for Pearson and 0.53 (Chip) to 0.76 (Tameka) for Spearman. Aggregating this single name choice by race shows Hispanic names have highest average performance (r: 0.64, ρ: 0.62), followed by White (r:0.63, ρ:0.61), Black (r:0.63, ρ:0.60), and Asian names (r:0.60, ρ:0.58). Notably, aggregating all single homogeneous names perform comparable to no-identifier baselines, suggesting that providing a uniform name across all students offers minimal benefit over no identifier (Table~\ref{tab:prompt_name_correlation}). However, diverse names consistently achieve the strongest correlations. Even when comparing classrooms with diverse names within single racial groups against classrooms with diverse names across all races, the fully diverse condition outperforms all race-specific diverse conditions across grades (Table~\ref{tab:all_correlations}), showing that cross-racial demographic diversity improved model's correlation.
We emphasize that these empirical results examine aggregate patterns across demographics; individual name results are never used to make claims about specific people, and all analyses report only aggregate statistics to characterize systematic model behavior rather than individual-level outcomes.
\begin{table}[h!]
\centering
\small
\resizebox{1\columnwidth}{!}{%
\begin{tabular}{lccccccc}
\toprule
\textbf{Prompt} & \multicolumn{3}{c}{\textbf{(r, PC) $\uparrow$}} & \multicolumn{3}{c}{\textbf{($\rho$, SC) $\uparrow$}} \\
\cmidrule(lr){2-4} \cmidrule(lr){5-7}
 & \textbf{G4} & \textbf{G8} & \textbf{G12} & \textbf{G4} & \textbf{G8} & \textbf{G12} \\
\midrule
No identifier & 0.62 & 0.57 & 0.67 & 0.61 & 0.55 & 0.66 \\ \bottomrule
Std. IDs & 0.66 & 0.60 & 0.72 & 0.64 & 0.58 & 0.72 \\ \bottomrule
Single & 0.61 & 0.59 & 0.68 & 0.60 & 0.57 & 0.65 \\
Diverse & \textcolor{black!70!black}{\textbf{0.72}} & \textcolor{black!70!black}{\textbf{0.66}} & \textcolor{black!70!black}{\textbf{0.74}} & \textcolor{black!70!black}{\textbf{0.71}} &\textcolor{black!70!black}{\textbf{0.64}} & \textcolor{black!70!black}{\textbf{0.74}} \\
\bottomrule
\end{tabular}
}
\caption{Correlations on prompt approaches for Gemma-2-9b across grades. Prompt approaches: No Name (no student identifier), Student IDs, Single Name, Diverse Names. All values statistically significant with p<0.01.}
\label{tab:prompt_name_correlation}
\end{table}


We also observe simulated performance correlates well with real world gender performance patterns (r: 0.69 to 0.74 and ρ: 0.69 to 0.79) in Table \ref{tab:demographic_correlations} and show similar correlation strengths. The differences between Pearson and Spearman correlations are minimal, suggesting that the relationship between simulated and real performance is both monotonic and approximately linear for gender. We assess generalization beyond the NAEP math
dataset and evaluate our pipeline on the EEDI \footnote{https://www.eedi.com}
dataset of UK middle school math questions \citep{heyueya2024psychometricalignmentcapturinghuman}. The
results reported in Appendix B.2, Table \ref{tab:method_comparison} show
our method of simulating students marginally exceeds supervised baselines. While not yet a complete replacement for human data, these results suggest that simulated responses have psychometric signal to inform item design and practical utility for pre-pilot screening of items.

\subsection{IRT Estimates Reflect Skill Profiles and Predict Real World Difficulty} 
\begin{figure}
    \centering
    \includegraphics[width=1\linewidth]{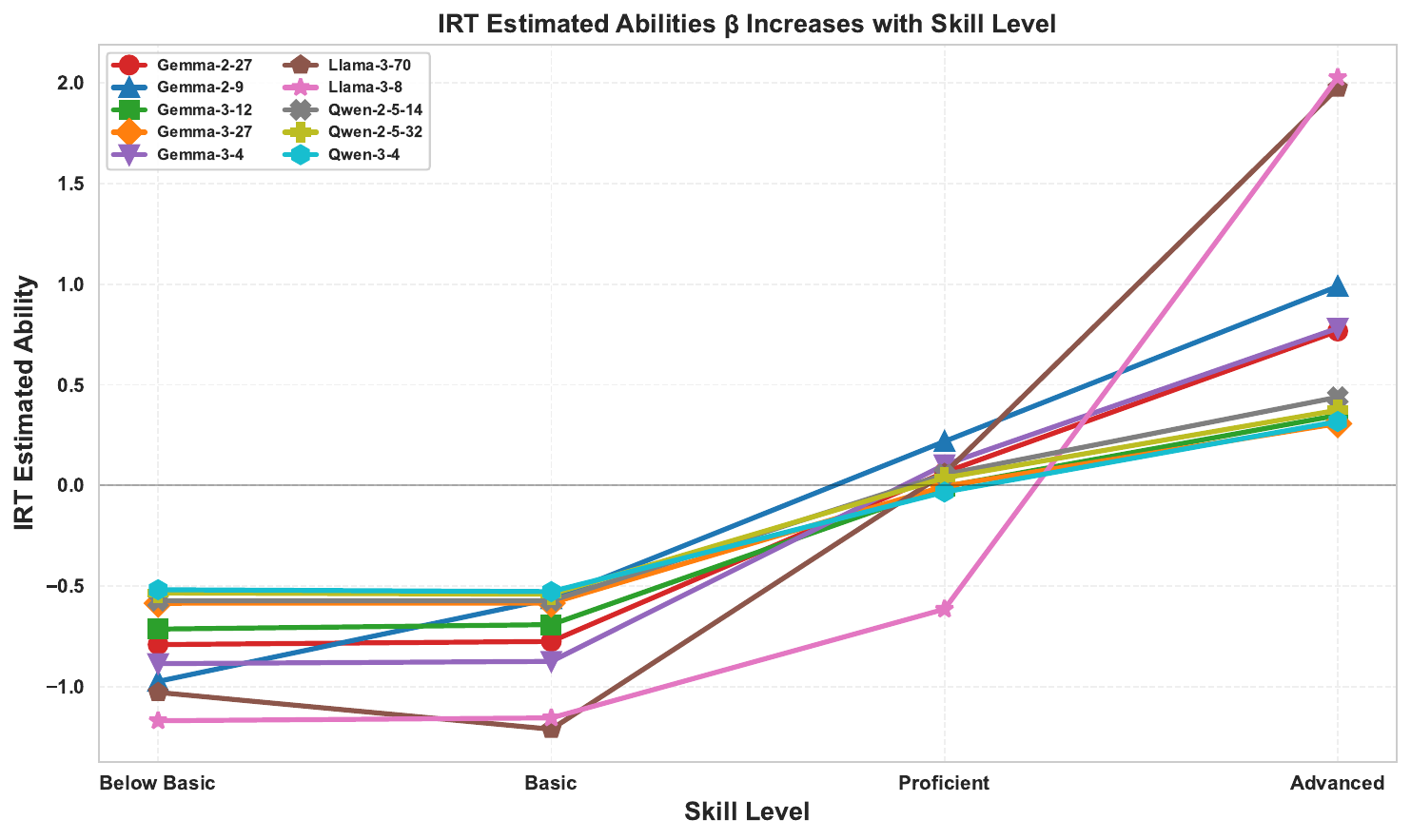}
   \caption{IRT group ability estimates for different LLMs across different skill levels.}
    \label{fig:irt_abilities}
\end{figure}

\textbf{Simulated students performance align with expected skill profiles.} The generated responses fitted to IRT models to extract difficulty parameters, provide a statistical framework for modeling how student group ability and item characteristics interact to produce responses. The results in Figure~\ref{fig:irt_abilities} show a monotonic increase from a Below Basic skill level to an Advanced skill level, although there seems to be a more sharper increase from the Basic to Proficient. This increase could be as result of the description of these skill levels, where Basic students are defined as student struggling when problems become more complex compared to the Proficient student  whose proficiency allows them to adapt and apply mathematical concepts to different problems irrespective of their complexity. We observe a slightly lower increase from the Below Basic to Basic, compared to the Proficient to Advanced, which may be attributed to an attenuated skill level abilities described in the prompt for these lower level skills lending to low mastery compared to higher skills.
 
We also measure the aggregated correctness rates of the group skill levels and present results in Figure~\ref{fig:correctness_rates} which show a consistent trend with the real world where lower skill levels on average have lower accuracy rates and this rate increases as we move from the below basic to advanced students.
In short, the simulated students' answers do align with their scripted skill levels, but the fidelity of that varies by LLM.
\noindent While we lack ground-truth data to evaluate whether LLM-generated reasoning/explanations for selecting correct answers are `human-like,' a manual analysis of these LLM generated explanations shows clear differences in certainty across skill levels. For lower-level skill levels, we observe LLMs tend to use hedging language that shows uncertainty \cite{Alseweed2005OvercomingUW}, including phrases like `I think/I guess,' or explicit admissions of arbitrary choices such as `I just picked 15 because it seemed like it could fit', while higher proficient students are more confident in  their reasoning for the answer selected (Tables~\ref{tab:sample_answer_1} to ~\ref{tab:sample_answer_3}).

\noindent \textbf{IRT estimated difficulties can predict real world difficulties.}
We examine if simulations capture the deeper psychometric structure needed for meaningful difficulty estimation. If our IRT-estimated difficulties only reflects simulation artifacts rather than true item properties, they would fail to predict how real students perform. We see in Table~\ref{tab:full_results} that the IRT difficulty parameters derived from simulated student responses show moderate to strong predictive validity for real world outcomes.  AUC specifically measures the models' understanding  of relative difficulty (which items are harder than others). The Gemma models achieved AUC scores ranging from 0.77 to 0.90 across different grade levels, indicating that simulated difficulty estimates can meaningfully discriminate between items that real students find easy versus hard. Thus our pipeline is able to generate responses that capture meaningful aspects of item difficulty that translate to real student performance, though the relationship varies by model and grade. We observe that model size does not guarantee better difficulty discrimination ability. 
\subsection{Simulated Students Do Not Reliably Predict Distractor Selection}
While our simulations predict whether students answer correctly, a natural follow-up is whether they predict which incorrect answer students select. Figure \ref{fig:distractors} reports distractor match rates, that is, the percentage of items where the most popular incorrect answer among simulated students matches the most popular incorrect answer among real students. Match rates across models range from approximately 31–47\%, only modestly above chance. Consistent with our content-area findings, Table \ref{tab:distractor_match} shows distractor match results are mostly highest on Measurement problems, and lowest on Algebra and Grade 8 items. This suggests an important boundary condition: our pipeline is a reliable predictor of item difficulty but is not yet a reliable model of error type. Future work should investigate whether incorporating student knowledge tracing or distractor-aware prompting can close this gap.

\section{Conclusion}
We show that open-source LLMs can simulate students to estimate math question difficulty, achieving correlations as high as 0.75, 0.76, and 0.82 for grades 4, 8, and 12, respectively, outperforming baseline text-based and direct prompting approaches. Our simulations differentiate student skill levels (Below Basic, Basic, Proficient, Advanced) in ways that align with psychometric expectations for predicting real student outcomes. We find that diverse names improve correlations over no names or homogeneous names, though the underlying mechanism requires further investigation. Content-area analysis reveals simulations excel at Measurement problems, but struggle with Algebra. Counterintuitively, most LLMs with relatively weaker mathematical abilities produce better student simulations than more mathematically proficient models, suggesting models should experience difficulty to authentically simulate struggling students. While these simulated students' computational costs are higher than direct methods (4-48 GPU-hours), this produces psychometrically valid difficulty estimates suitable for preliminary item screening before expensive human piloting.
\section*{Limitations}

\paragraph{Distractor Answer Selection}
We included new analysis in Figure \ref{fig:distractors} and Table~\ref{tab:distractor_match} examining whether simulated students select the same incorrect answers (distractors) as real students. Figure \ref{fig:distractors} reports the percentage of exact match on the top-1 distractor answers selected by our simulated students and the real world students, to understand if the two populations share a most-popular distractor. Table \ref{tab:distractor_match} measures the distractor match rate,  the percentage of items where the most popular incorrect answer among simulated students matches the most popular incorrect answer among real students (random chance: ~25\% for 4-choice MCQ and ~20\% for 5-choice MCQ).
We observe relatively low matches with all models, which suggests that these models are useful predictors of whether students will get a question right or wrong, but they aren’t very good models of *how* students get questions wrong (i.e. which wrong answer they will pick). We see a similar trend with the content area correlation results which shows the real world distractor answer selection has relatively more distractor alignment on Measurement problems for Gemma 2-9b-it, and matches are weakest on Algebra, reflecting greater diversity in algebraic reasoning errors. However, as we focus the scope of this work to our primary goal of using the correctness rates, so future work could look at developing better models of how students get questions wrong.

\paragraph{LLM Limitations}
Our experiments relied on open-source language models, which may not reflect the upper bounds of performance achievable with larger, proprietary models such as GPT-4. It is possible that such models, although more expensive would provide more accurate simulations of student behavior, potentially narrowing the performance gap between direct and generative prompting strategies. Expanding the model pool with varying model families and architecture can also provide more robust conclusions and generalization of our findings. We deliberately focus on open-source models to enable reproducibility and cost-effective deployment. While proprietary models may achieve different results, our goal is demonstrating practical screening tools accessible to educational practitioners without API costs.

\paragraph{Limited Data size and question types} 
We evaluated model-generated responses on 631 multiple-choice questions for Grade 4, 8 and 12. While these cover a range of difficulty levels and content areas, the size and scope of the questions remain constrained for math Multiple choice questions. We focus mostly on MCQs as they provide a clear correctness criteria, so we can validate these computed IRT difficulty estimates, more effectively. We acknowledge that the dataset that we use in this work is limited in size (number of problems) and from a single source (NAEP), and highlight some of the uniquely enabling strengths that this data source offers, and why we chose to use it for our study. (1) Because our goal is to measure how well LLMs can predict the difficulty of math problems for real students, we felt that the single most important quality of the dataset was the quality of the associated real-world test performance statistics. In this regard, we believe the NAEP-reported performance statistics associated with each of the 631 math problems is particularly invaluable, and not easily obtainable from other sources. In particular, the NAEP statistics are collected as part of a U.S. congressionally mandated effort to assess educational achievement across public and private school districts across the U.S. The NAEP tests are administered to be nationally representative. Thus the resulting performance statistics reflect a large pool of test-takers, specifically sampled to be representative of the broader U.S. student population, in terms of gender, race and ethnicity, socioeconomic status, geographic location, and public/private school status, among others. This gives us confidence that the associated student test accuracy rates associated with each test item is broadly representative of the U.S. population, and is much less subject to the kinds of selection population biases that would be unavoidable with smaller-scale, privately collected data. We acknowledge the limitation that the U.S. population is not reflective of the broader world-wide population; however, we believe it is a strength of this data that the corresponding population is well-defined and well-represented therein, even though it is not universal.
Future work could study simulations for non-mathematics subjects (e.g., reading and science) where misconception patterns differ substantially; evaluating open-ended and constructed-response formats beyond multiple-choice; and developing richer student persona models that incorporate socioeconomic status, multilingual background, and prior knowledge state to improve simulation realism and fairness.

\paragraph{Limited diversity in demographics, grade and class size experiments}
We simulated student personas using 48 distinct first names distributed across four racial/ethnic groups (Black, Asian, Hispanic, and White) and two  genders. While this offers a starting point for exploring demographic variation, it does not capture the full richness and intersectionality of real classrooms. Broader name sets, additional identity dimensions (e.g., socioeconomic status, multilingual background), and intersectional profiles could allow for a more fine-grained analysis of item performance and fairness. 
We recognize that our use of demographic name proxies may activate stereotypes encoded in model training data. While our aggregate results show diverse names improve correlation, we cannot determine whether this reflects: (1) stereotype-based performance simulation (models enacting biased assumptions about name-associated demographics), (2) increased prompt specificity providing better grounding, or (3) other mechanisms. Individual name variance (SD=0.059-0.090) could suggest potential stereotype activation, though we emphasize all analyses report only aggregate statistics across demographic categories, never making individual-level predictions. Future work can investigate whether observed performance patterns reflect genuine diversity benefits or problematic stereotype reinforcement.

\noindent Our simulations were also constrained to Grade 4, Grade 8 and Grade 12 students, however, student behavior and response patterns may differ in early primary or upper high school levels. Extending the approach to other grades could uncover new insights or limitations.
For each test item, we simulated responses from between 100-300 students. Although we observed improved correlation with real-world data as sample size increased, we limited our simulations to manage resource costs. Larger sample sizes may offer more stable performance estimates and more realistic modeling of population-level variance, but at a greater computational cost.

\paragraph{Data Leakage}
We acknowledge that NAEP items are publicly accessible and may appear in LLM training data. However, this does not invalidate our approach because: (1), our task is fundamentally different from solving problems: we evaluate whether simulated response distributions match real student performance rates, not whether the LLM can identify the correct answer, (2), direct prompting for performance statistics produce near-zero correlations across all models (Tables \ref{tab:direct_results} – \ref{tab:averaged_results}), including GPT-4o, suggesting that if models had memorized NAEP statistics, direct estimation would succeed, (3) our simulation method outperforms direct prompting, suggesting that the correlation comes from the simulation structure rather than retrieval, (4) NAEP item-level student performance statistics are stored and reported separately from item texts, making co-occurrence in training corpora unlikely, without further efforts. Thus, even potential exposure to questions doesn't compromise our evaluation of difficulty estimation.

\section*{Ethics Statement}
In this study, we simulate student responses using a large language model (LLM) and vary the first names of hypothetical students—selecting names statistically associated with different genders and racial/ethnic groups. We acknowledge that inferring or assigning demographic identities based on first names is an inherently imperfect and sensitive approach, one that carries the risk of overgeneralization or reinforcement of stereotypes. A first name is at best a loose proxy for a demographic group, and relying on names can inadvertently evoke stereotypical assumptions if not handled carefully. To mitigate these concerns, we employ first-name variations purely as a controlled variable in a bias audit context, ensuring that any observed performance differences are attributed to the model’s behavior or potential biases in the content rather than presumed traits of any real group. We emphasize that these are empirical observations and that they correlate with improved alignment in our experiments.

We further recognize the broader risk that large language models may reproduce or amplify societal biases present in their training data. In our simulations, the model’s outputs could reflect such historical biases or stereotypes—for example, it might yield systematically different responses or difficulty assessments for different name conditions, echoing real-world disparities. Our intent, however, is to leverage these controlled simulations to identify and understand potential inequities, not to perpetuate them. 

\section*{Acknowlegements}
We thank the anonymous reviewers for their constructive feedback. We are also grateful to Haozhe An, Kwesi Cobbina, Neha Srikanth, Nishant Balepur, Rupak Sarkar, and the rest of rlab for their valuable advice and support throughout this project.

\bibliography{anthology,custom}

\appendix
\section{Appendix}

\subsection{Data Details}
\label{sec:data}

\begin{figure}
    \centering
    \includegraphics[width=1\linewidth]{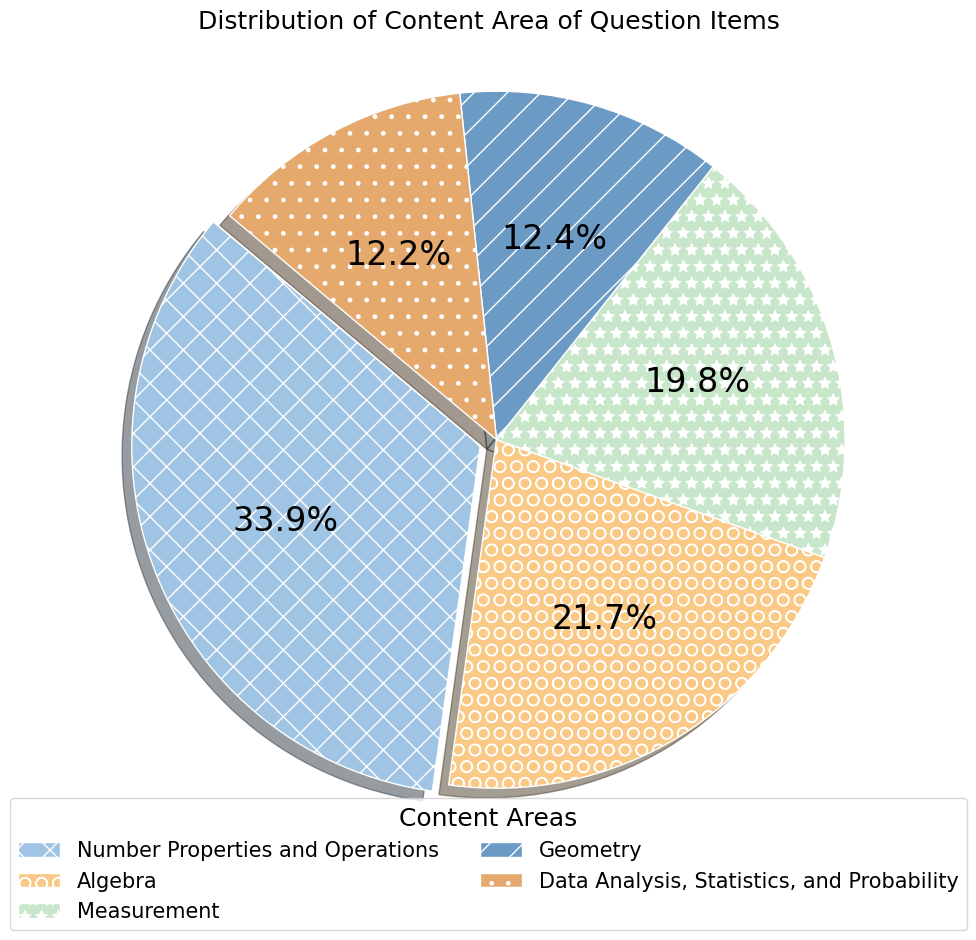}
    \caption{Distribution of content areas being tested in the dataset.}
    \label{fig:question_stats}
\end{figure}
We present two examples of question texts  from our collected data. 
\\ 
\\
\noindent\fbox{%
    \begin{minipage}{1\columnwidth}
     \scriptsize
    \textbf{Example 1: }
    
    \textit{\textbf{Sebastian}} is making lemonade. His recipe requires 750 grams of sugar to make 20 liters of lemonade. Sebastian wants to make 12 liters of lemonade. How many grams of sugar does Sebastian need to maintain the same ratio of sugar to lemonade as in his recipe?
\\
\\
    \textbf{Example 2: }
    \textit{\textbf{Ms. Thierry}} and 3 friends ate dinner at a restaurant. The bill was \$67. In addition, they left a \$13 tip. Approximately what percent of the total bill did they leave as a tip?
    \end{minipage}
    } 

\begin{table}[h!]
\centering
\resizebox{1\linewidth}{!}{ 
    \begin{tabular}{lrrrr}
        \toprule
        \textbf{Difficulty} & \textbf{Total \#} & \textbf{ Grade 4}& \textbf{Grade 8}& \textbf{Grade 12}\\
        \midrule
       Easy                      & 254               & 102              & 108              & 44                \\
Medium                    & 217               & 76               & 106              & 35                \\
Hard                      & 160               & 50               & 68               & 42                \\
\hline
\textbf{Total}            & \textbf{631}      & \textbf{228}     & \textbf{282}     & \textbf{121}      \\
        \bottomrule
    \end{tabular}
}
\caption{Breakdown of question difficulty by grade.}
\label{tab:difficulty_gender_breakdown}
\end{table}
\subsection{Prompts}
\label{sec:Prompts}
\begin{prompt}[title={Prompt \thetcbcounter: Baseline-Knowledge Prompt}, label=prompt:knowledge_prompt]
\scriptsize
**Task**: \\
You are an expert problem solver. Solve step by step the following math word problems. Only respond with the letter of
the correct answer. Prefix your final answer with Answer Key: [letter]".
\\

\end{prompt}

\begin{prompt}[title={Prompt \thetcbcounter: Baseline-Direct Simulation Prompt}, label=prompt:direct_prompt]
\scriptsize
**Task**: \\
You are an expert in predicting student performance. Given this math word problem written for \{grade\}th-grade students, estimate the percentage of students at this grade level who will answer the question correctly. Your prediction should be based on factors such as problem difficulty and cognitive load at this grade level.
Prefix your final answer with "Percentage Correct: [percentage]".
                
\end{prompt}

\begin{prompt}[title={Prompt \thetcbcounter: Student Simulation Prompt}, label=prompt:student_prompt]
\scriptsize
**Task**: \\
You are a \{skill level\} student in the \{grade\}th grade, given the task to answer a math word problem question on \{content area of problem\}, taking into account the difficulty of this question. \{Definition of skill level continues\}.

In all your responses, you have to completely forget that you are an AI model, but rather this \{skill level\} student, and completely simulate yourself as one.
\end{prompt}



\begin{prompt}[title={Prompt \thetcbcounter: Demographic Student Simulation Prompt}, label=prompt:demographic_prompt]
\scriptsize
**Task**: \\
You are [NAME], a student in the \{grade\}th grade, given the task to answer a math word problem question on \{content area of problem\}, taking into account the difficulty of this question. \{Definition of skill level continues\}.

In all your responses, you have to completely forget that you are an AI model, but rather this student  [NAME], and completely simulate yourself as one.
        
\end{prompt}

\begin{prompt}[title={Prompt \thetcbcounter: Student ID Simulation Prompt}, label=prompt:ids_prompt]
\scriptsize
**Task**: \\
You are [STDID], a student in the \{grade\}th grade, given the task to answer a math word problem question on \{content area of problem\}, taking into account the difficulty of this question. \{Definition of skill level continues\}.

In all your responses, you have to completely forget that you are an AI model, but rather this student [STDID], and completely simulate yourself as one.
        
\end{prompt}

\subsection{Names}
\label{sec:appendix_first_names}

The names used in our experiments are listed. 

\paragraph{Asian female names}
Syeda, Thuy, Eun, Ngoc, Inaaya, Priya

\paragraph{Asian male names}
Aryan, Vihaan, Armaan, Quang, Trung, Chang

\paragraph{Black female names}
Latoya, Lashelle, Imani, Shante, Tameka, Nichelle

\paragraph{Black male names}
Malik, Leroy, Darius, Tyrone, Rashaun, Cedric

\paragraph{Hispanic female names}
Alejandra, Xiomara, Mariela, Migdalia, Marisol, Julissa

\paragraph{Hispanic male names}
Lazaro, Osvaldo, Alejandro, Jairo, Heriberto, Guillermo

\paragraph{White female names}
Susan, Courtney, Kimberly, Heather, Barbara, Molly

\paragraph{White male names}
Charles, Roger, Wilbur, Hank, Chip, Hunter
\\


\subsection{Student IDs Used}
\label{sec:UIDs}

We generated alphanumeric identifiers 
following the format \texttt{STU} followed by a 6-digit zero-padded number (e.g., \texttt{STU000001}, \texttt{STU600298}). IDs were randomly generated to avoid 
any systematic patterns, with randomly sampled from 000000-99999 
and prepended with \texttt{STU} to create the identifier.


\section{Additional Experiment}
\subsection{Experimental Setup Details}
\label{sec:appendix_experiments}
\paragraph{Experimental Design Summary}
\label{sec:appendix_experiments_}
We formalize our simulation as a function with the parameters:

\begin{equation}
\text{Correlation} = f(\mathcal{M}, \mathcal{S}, \mathcal{I}, N, \mathcal{E})
\end{equation}

where:\begin{itemize}
    \item $\mathcal{M}$: \textbf{Model selection} -- which LLM(s) is used
    \item $\mathcal{S}$: \textbf{Skill distribution} -- proportion of Below Basic, Basic, Proficient, Advanced students
    \item $\mathcal{I}$: \textbf{Identifier strategy} -- how students are identified (No Identifier, Student IDs, Single Name, Diverse Names)
    \item $N$: \textbf{Classroom size} -- number of simulated students per question
    \item $\mathcal{E}$: \textbf{Estimation method} -- how difficulty is computed (Direct prompt, Simulated responses + IRT)
\end{itemize}
Table~\ref{tab:experiment_summary} summarizes all experimental configurations and their corresponding results.

\begin{table*}[h]
    \centering
    \resizebox{\textwidth}{!}{
    \small
    \begin{tabular}{lp{4cm}p{3cm}p{2cm}}
        \toprule
        \textbf{Experiment} & \textbf{Purpose} & \textbf{Key Parameters} & \textbf{Models} \\
        \midrule
        Baseline: QDET & Establish text-based difficulty estimation baseline & Method: Word2Vec, BERT, DistilBERT & BERT, DistilBERT\\
        \midrule
        Baseline: Direct Prompting & Establish direct LLM prompting baseline & Method: Greedy \& averaged predictions & All 10 models \\
        \bottomrule
        Core Simulation & Evaluate simulation fidelity across models & $\mathcal{M}$: 10 models; $\mathcal{I}$: No Name; $N$: 300 & All 10 models \\
        \midrule
        Model Ensembling & Test if combining models improves correlation & $\mathcal{E}$: Averaged vs. Weighted ensemble & Averaged (all 10), Weighted (top 3) \\
        \midrule
        Class Size Variation & Determine optimal classroom size & $N$: 50, 100, 300 students & Gemma-2-9b \\
        \midrule
        Identifier Strategy & Compare naming/ID approaches & $\mathcal{I}$: None, IDs, Single Name, Diverse Names & Gemma-2-9b \\
        \midrule
        Within-Category Diversity & Test racial homogeneity vs. diversity & $\mathcal{I}$: Single-race diverse vs. all-race diverse & Gemma-2-9b \\
        \midrule
        Gender Correlation & Validate against gender-specific performance & $\mathcal{I}$: Diverse Names (gender-specific analysis) & Gemma-2-9b \\
        \midrule
        Content Area Analysis & Identify domain-specific strengths/weaknesses & Stratify by: Algebra, Geometry, Measurement, etc. & All 10 models \\
        \midrule
        Skill Level Alignment & Verify IRT ability estimates match skill levels trend expectations& $\mathcal{S}$: NAEP 4-level distribution & All 10 models \\
        \bottomrule
    \end{tabular}
    }
    \caption{Summary of experimental configurations. All simulations use NAEP skill distribution (25\% Below Basic, 35\% Basic, 25\% Proficient, 15\% Advanced) unless noted otherwise.}
    \label{tab:experiment_summary}
\end{table*}

\paragraph{Computational Cost} QDET baselines require less than 1 minute for transformer inference across all 631 items on A6000. While our simulation approach using open-source models requires approximately 4-5 GPU-hours on A6000 GPUs for models in the 8-12B parameter range to generate 300 simulated student responses across all 631 items. Larger models (e.g., 70B parameter) would require proportionally more time, though we found the correlation strengths are not comparable with model scale. For example the Gemma-2-9B needs 4.34 GPU-hours on 1 A 6000, for all 631 items for 300 students and Llama-3.3-70B requires up to 48 hours with 4 A6000s for 300 student simulation.

\paragraph{Terms of use for each model}
We carefully follow the guidelines per the terms of usage described by the model authors or company
\begin{itemize}
    \item Gemma2: \url{https://github.com/google-deepmind/gemma/blob/main/LICENSE}
    \item Gemma-3: \url{https://ai.google.dev/gemma/terms}
    \item Llama3: \url{https://llama.meta.com/llama3/license/}
    \item Qwen2.5/3: \url{https://github.com/QwenLM/Qwen/blob/main/LICENSE} 
\end{itemize}
\textbf{Licenses} The NAEP data is used under the  MIT\footnote{\textcolor[HTML]{000099}{https://opensource.org/license/MIT}} and CC-BY\footnote{\textcolor[HTML]{000099}{https://creativecommons.org/licenses/by/4.0/}} licenses. 

\subsection{Additional  Results}
\paragraph{Generalizations to EEDI Math Datasets}
To test generalizability beyond NAEP, we evaluate on the EEDI dataset, a real-world student response dataset of math MCQs. We select a random subset of 150 question items from this EEDI dataset and compute ground truth from the (1/0) correctness rates of the individual student responses. We use \citet{Feng}'s method, with a  5-fold cross-validation and  train a Longformer model with GPT-4o-generated reasoning augmentation for 300 epochs at their specified learning rates. We generated 300 simulated student responses using Gemma 2-9b-it, for all 150 items. We focus mainly on ranking-based metrics as our simulated and empirical difficulties are on different scales. Our downstream use of difficulty prediction rely on relative ordering rather than exact values. We report MATCH (measures  percentage of questions where the predicted relative difficulty ranking matches the ground truth) below.  We note that a higher MATCH score does not necessarily imply higher correlations, as MATCH only captures pairwise direction (whether question A is harder than B), while correlations measure the overall consistency of the predicted ranking.

\begin{table*}
\centering
\resizebox{1\columnwidth}{!}{%
\begin{tabular}{lccc}
\toprule
\textbf{Method} &\textbf{MATCH $\uparrow$} & \textbf{Pearson $\uparrow$} & \textbf{Spearman $\uparrow$} \\
\midrule
FT&0.55$_{\pm0.013}$& 0.20$_{\pm0.0425}$& 0.15$_{\pm0.047}$\\
Feng et al.\ FT &0.58$_{\pm0.018}$& 0.28$_{\pm0.0935}$& 0.25$_{\pm0.0589}$\\
Our method (Simulated Students) &0.5927& 0.3105& 0.2655\\
\bottomrule
\end{tabular}
}
\caption{Comparison of methods by MATCH, Pearson, and Spearman correlation. Values shown as mean$_{\pm\text{std}}$. `FT' is Longformer fine-tuning without GPT-4o reasoning augmentation; `Feng et al. FT' augments FT with GPT-4o reasoning. Both baselines are trained with 5-fold cross-validation following \citet{Feng} . Our simulation method requires no training, so we report single-run results on all 150 items.}
\label{tab:method_comparison}
\end{table*}
\begin{figure*}
    \centering
    \includegraphics[width=1\linewidth]{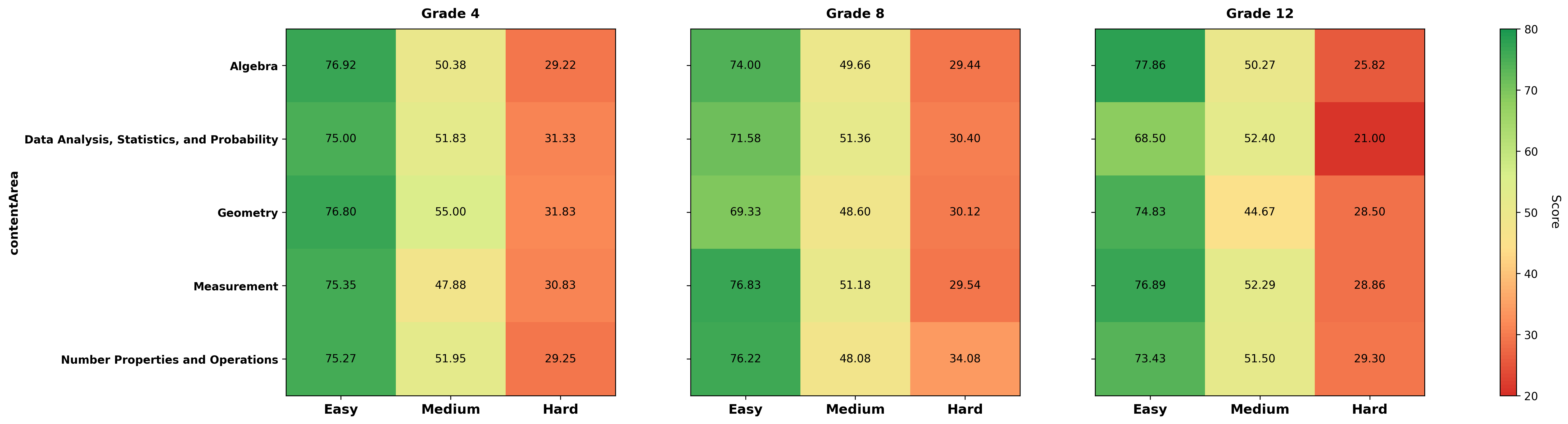}
    \caption{Average Percentage Correct Score by NAEP Assigned Difficulty at a grade and content area breakdown. }
    \label{fig:human_correct}
\end{figure*}

\begin{figure*}
    \centering
    \includegraphics[width=1\linewidth]{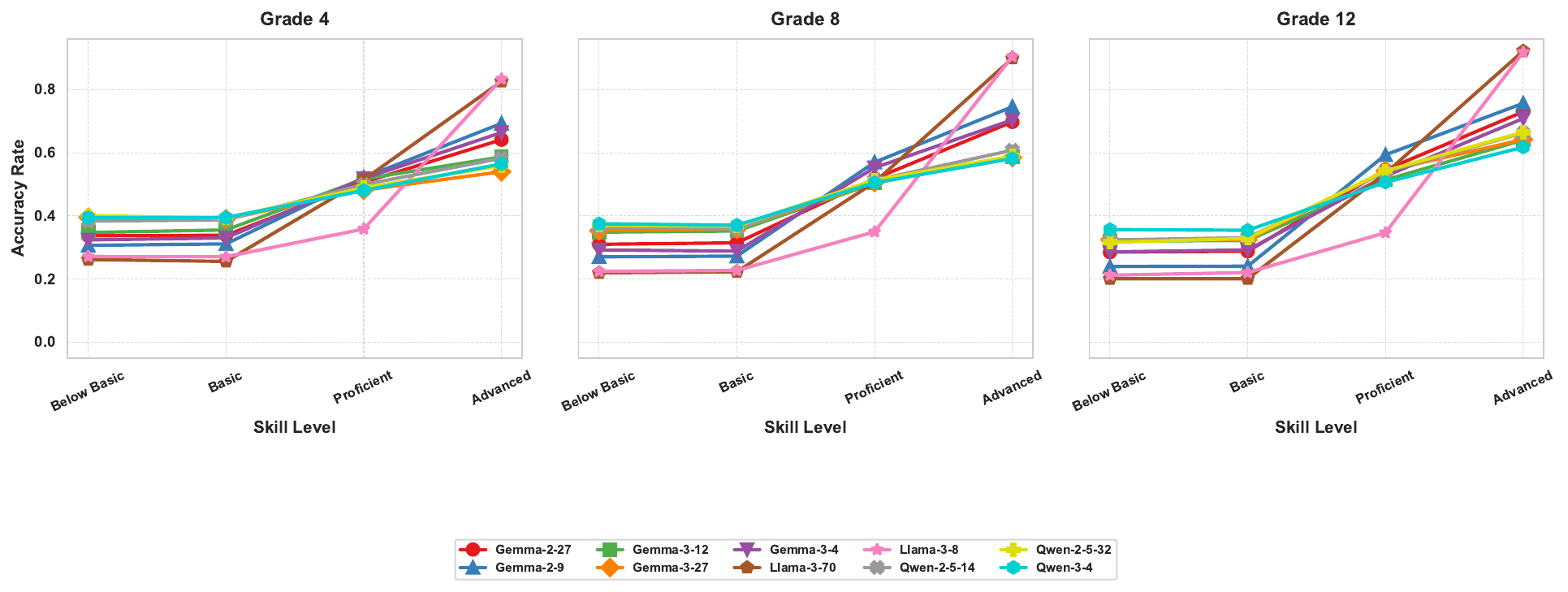}
    \caption{Average Accuracy rates for simulated students of different skill levels across grades, 4, 8 and 12.  
    }
    \label{fig:correctness_rates}
\end{figure*}

\begin{table*}[h!]
\centering
\small
\begin{tabular}{lccc}
\toprule
\textbf{Approach} & \textbf{Grade 4} & \textbf{Grade 8} & \textbf{Grade 12} \\
\midrule
Word2Vec + RF & 0.36 & -0.04 & 0.30 \\
BERT & 0.05 & 0.03 & 0.21 \\
DistillBERT & 0.04 & 0.01 & 0.17 \\
\bottomrule
\end{tabular}
\caption{Comparison of QDET performance across different grade levels.}
\label{tab:approach_results}
\end{table*}
\begin{table*}[h!]
\centering
\small
\begin{tabular}{lccccccc}
\toprule
\textbf{Gender} & \multicolumn{3}{c}{\textbf{(r, PC) $\uparrow$}} & \multicolumn{3}{c}{\textbf{($\rho$, SC) $\uparrow$}} \\
\cmidrule(lr){2-4} \cmidrule(lr){5-7}
 & \textbf{G4} & \textbf{G8} & \textbf{G12} & \textbf{G4} & \textbf{G8} & \textbf{G12} \\
\midrule
Females & 0.69 & \textcolor{black!70!black}{\textbf{0.73}}  & 0.74 & 0.69 & \textcolor{black!70!black}{\textbf{0.75}} & 0.77 \\
Males & \textcolor{black!70!black}{\textbf{0.70}} & 0.71 & \textcolor{black!70!black}{\textbf{0.74}} & \textcolor{black!70!black}{\textbf{0.70}} & \textcolor{black!70!black}{\textbf{0.75}} & \textcolor{black!70!black}{\textbf{0.79}} \\
\bottomrule
\end{tabular}
\caption{Correlation results with real world gender demographic groups based on diverse names for Gemma-2-9b across grades. All values statistically significant with p<0.01.}
\label{tab:demographic_correlations}
\end{table*}
\begin{table*}
\centering
\small
\resizebox{\textwidth}{!}{%
\begin{tabular}{lcccccccccccc}
\toprule
\multirow{2}{*}{\textbf{Model}} & \multicolumn{3}{c}{\textbf{(r, PC) $\uparrow$}} & \multicolumn{3}{c}{\textbf{($\rho$, SC) $\uparrow$}} & \multicolumn{3}{c}{\textbf{$R^2$ $\uparrow$}} & \multicolumn{3}{c}{\textbf{AUC $\uparrow$}} \\
\cmidrule(lr){2-4} \cmidrule(lr){5-7} \cmidrule(lr){8-10} \cmidrule(lr){11-13}
 & \textbf{G4} & \textbf{G8} & \textbf{G12} & \textbf{G4} & \textbf{G8} & \textbf{G12} & \textbf{G4} & \textbf{G8} & \textbf{G12} & \textbf{G4} & \textbf{G8} & \textbf{G12}\\
\midrule
Gemma-2-9b  & \textbf{0.69} {\scriptsize [0.61, 0.74]} & 0.61 {\scriptsize [0.52, 0.67]} & 0.74 {\scriptsize [0.63, 0.80]} & \textbf{0.69} {\scriptsize [0.59, 0.74]} & 0.59 {\scriptsize [0.49, 0.66]} & 0.73 {\scriptsize [0.61, 0.81]} & \textbf{0.46} & 0.36 & 0.53 & 0.84 & 0.78 & 0.84\\
Gemma-2-27b & \textbf{0.69} {\scriptsize [0.62, 0.75]} & \textbf{0.68} {\scriptsize [0.62, 0.74]} & \textbf{0.78} {\scriptsize [0.71, 0.84]} & \textbf{0.68} {\scriptsize [0.61, 0.75]} & \textbf{0.67} {\scriptsize [0.59, 0.73]} & \textbf{0.78} {\scriptsize [0.70, 0.83]} & \textbf{0.48} & \textbf{0.46} & \textbf{0.61} & \textbf{0.85} & \textbf{0.83} & \textbf{0.90}\\
\midrule
Gemma-3-4b  & 0.58 {\scriptsize [0.49, 0.66]} & 0.57 {\scriptsize [0.49, 0.64]} & 0.68 {\scriptsize [0.58, 0.76]} & 0.57 {\scriptsize [0.47, 0.65]} & 0.54 {\scriptsize [0.45, 0.62]} & 0.66 {\scriptsize [0.54, 0.76]} & 0.33 & 0.32 & 0.46 & 0.77 & 0.77 & 0.83\\
Gemma-3-12b & 0.66 {\scriptsize [0.57, 0.72]} & \textbf{0.70} {\scriptsize [0.64, 0.76]} & 0.76 {\scriptsize [0.67, 0.81]} & \textbf{0.67} {\scriptsize [0.56, 0.74]} & \textbf{0.70} {\scriptsize [0.62, 0.77]} & 0.75 {\scriptsize [0.65, 0.80]} & 0.42 & \textbf{0.50} & 0.55 & 0.82 & 0.82 & \textbf{0.88}\\
Gemma-3-27b & \textbf{0.70} {\scriptsize [0.62, 0.76]} & \textbf{0.72} {\scriptsize [0.66, 0.77]} & 0.75 {\scriptsize [0.67, 0.82]} & \textbf{0.74} {\scriptsize [0.66, 0.81]} & \textbf{0.72} {\scriptsize [0.65, 0.78]} & 0.75 {\scriptsize [0.66, 0.83]} & \textbf{0.49} & \textbf{0.52} & 0.56 & \textbf{0.88} & \textbf{0.85} & \textbf{0.89}\\
\midrule
Llama-3-8b    & 0.42 {\scriptsize [0.32, 0.52]} & 0.44 {\scriptsize [0.34, 0.53]} & 0.44 {\scriptsize [0.31, 0.58]} & 0.40 {\scriptsize [0.28, 0.51]} & 0.46 {\scriptsize [0.36, 0.55]} & 0.48 {\scriptsize [0.32, 0.62]} & 0.18 & 0.19 & 0.20 & 0.67 & 0.72 & 0.75\\
Llama-3-3-70b & 0.56 {\scriptsize [0.47, 0.65]} & 0.53 {\scriptsize [0.44, 0.60]} & 0.46 {\scriptsize [0.33, 0.59]} & 0.54 {\scriptsize [0.43, 0.64]} & 0.52 {\scriptsize [0.43, 0.60]} & 0.46 {\scriptsize [0.31, 0.58]} & 0.32 & 0.28 & 0.21 & 0.73 & 0.74 & 0.75\\
\midrule
Qwen2.5-14B & 0.59 {\scriptsize [0.50, 0.68]} & 0.61 {\scriptsize [0.52, 0.68]} & 0.58 {\scriptsize [0.47, 0.68]} & 0.61 {\scriptsize [0.51, 0.70]} & 0.61 {\scriptsize [0.53, 0.69]} & 0.57 {\scriptsize [0.43, 0.68]} & 0.35 & 0.37 & 0.34 & 0.80 & 0.79 & 0.80\\
Qwen2.5-32B & 0.60 {\scriptsize [0.51, 0.68]} & 0.62 {\scriptsize [0.55, 0.69]} & 0.51 {\scriptsize [0.40, 0.62]} & \textbf{0.64} {\scriptsize [0.54, 0.73]} & \textbf{0.65} {\scriptsize [0.56, 0.72]} & 0.50 {\scriptsize [0.37, 0.61]} & 0.36 & 0.39 & 0.27 & 0.82 & 0.81 & 0.76\\
\midrule
Qwen3-4b & 0.60 {\scriptsize [0.51, 0.68]} & 0.58 {\scriptsize [0.49, 0.66]} & 0.62 {\scriptsize [0.51, 0.72]} & \textbf{0.64} {\scriptsize [0.54, 0.73]} & 0.63 {\scriptsize [0.54, 0.71]} & 0.62 {\scriptsize [0.48, 0.73]} & 0.36 & 0.34 & 0.38 & 0.79 & 0.80 & 0.83\\
\bottomrule
\end{tabular}
}
\caption{Full Correlation, $R^2$, and AUC Results of Simulated Students by Model and Grade. All correlations are statistically significant at p < 0.01. 95\% CIs computed via 1000 bootstrap resamples. \textit{r} = Pearson correlation; $\rho$ = Spearman correlation; $R^2 = r^2$ (proportion of variance explained); AUC = Area Under ROC Curve.}
\label{tab:full_results_all}
\end{table*}

\begin{table*}[t]
\centering
\small
\begin{tabular}{lcccccc}
\toprule
 \textbf{Names Used} & \multicolumn{2}{c}{\textbf{Grade 4}} & \multicolumn{2}{c}{\textbf{Grade 8}} & \multicolumn{2}{c}{\textbf{Grade 12}} \\
 & {\textbf{(r, PC) $\uparrow$}} & {\textbf{($\rho$, SC) $\uparrow$}} & {\textbf{(r, PC) $\uparrow$}} & {\textbf{($\rho$, SC) $\uparrow$}} & {\textbf{(r, PC) $\uparrow$}} & {\textbf{($\rho$, SC) $\uparrow$}} \\
\midrule
Diverse Asian Names & 0.502$_{\pm0.051}$ & 0.474$_{\pm0.030}$ & 0.513$_{\pm0.018}$ & 0.509$_{\pm0.024}$ & 0.572$_{\pm0.037}$ & 0.562$_{\pm0.029}$ \\
 Diverse Black Names & 0.536$_{\pm0.052}$ & 0.495$_{\pm0.023}$ & 0.513$_{\pm0.026}$ & 0.514$_{\pm0.021}$ & 0.585$_{\pm0.028}$ & 0.559$_{\pm0.019}$ \\
 Diverse Hispanic Names & 0.503$_{\pm0.049}$ & 0.491$_{\pm0.049}$ & 0.519$_{\pm0.030}$ & 0.516$_{\pm0.031}$ & 0.580$_{\pm0.027}$ & 0.567$_{\pm0.029}$ \\
Diverse White Names& 0.575$_{\pm0.054}$ & 0.499$_{\pm0.022}$ & 0.514$_{\pm0.019}$& 0.487$_{\pm0.020}$& 0.622$_{\pm0.026}$ & 0.578$_{\pm0.030}$ \\ 
\midrule
 Diverse Names & \textbf{0.596}$_{\pm0.027}$ & \textbf{0.557}$_{\pm0.038}$ & \textbf{0.582}$_{\pm0.028}$& \textbf{0.554}$_{\pm0.034}$& \textbf{0.639}$_{\pm0.049}$ & \textbf{0.612}$_{\pm0.047}$ \\
\bottomrule
\end{tabular}
\caption{Correlations across different name conditions and grades with standard deviations. r = Pearson correlation, $\rho$ = Spearman correlation. Values shown as mean$_{\pm\text{SD}}$. We simulate smaller classroom sizes to ensure unique name assignment, with each student sampled from different skill levels. Within each race condition (Asian, Black, Hispanic, White), names are sampled to ensure diversity in gender (male/female) while maintaining racial homogeneity. For example, the Asian condition uses different Asian male and female names across students. The "Diverse Names" condition samples names across all four racial groups while maintaining gender diversity and unique skill-level representation. The lower correlations compared to Table~\ref{tab:prompt_name_correlation} reflect the reduced classroom size, yet the pattern persists: diverse names across racial groups  outperform racially homogeneous conditions, showing that cross-racial diversity improves correlations} 
\label{tab:all_correlations}
\end{table*}

\begin{figure*}

    \centering
    \includegraphics[width=1\linewidth]{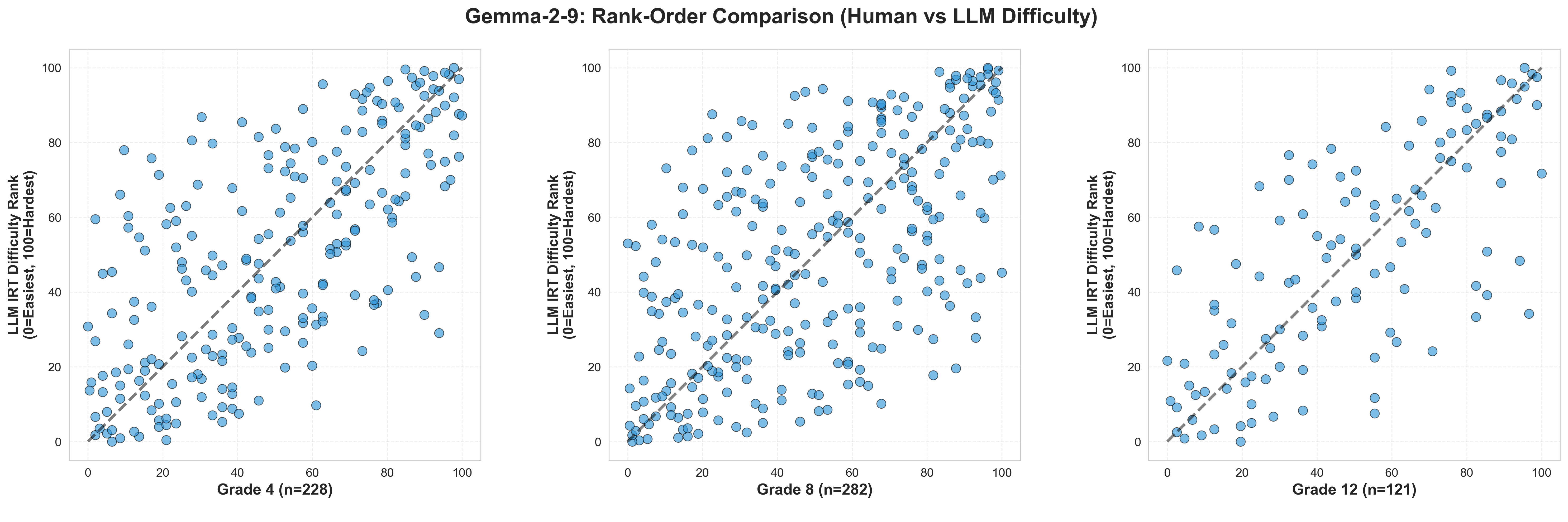}
    \label{fig:scatter_plots}
\end{figure*}
\begin{figure*}
    \centering
    \includegraphics[width=0.85\linewidth]{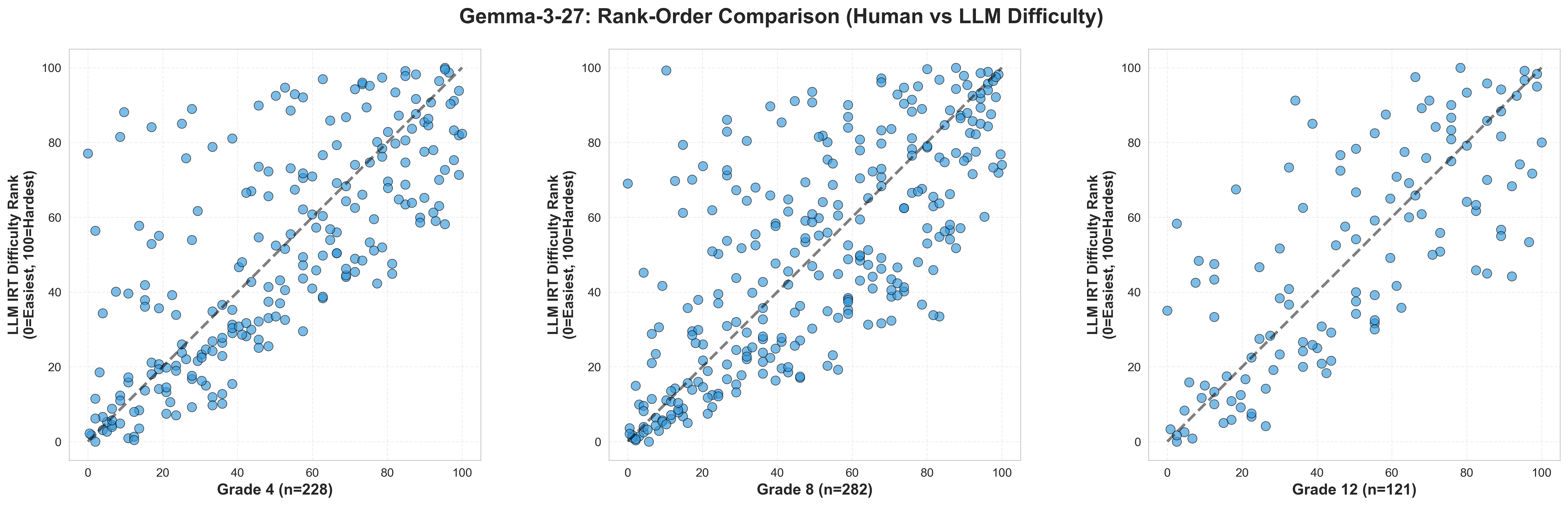}
    \label{fig:placeholder}
\end{figure*}
\begin{figure*}
    \centering
    \includegraphics[width=0.85\linewidth]{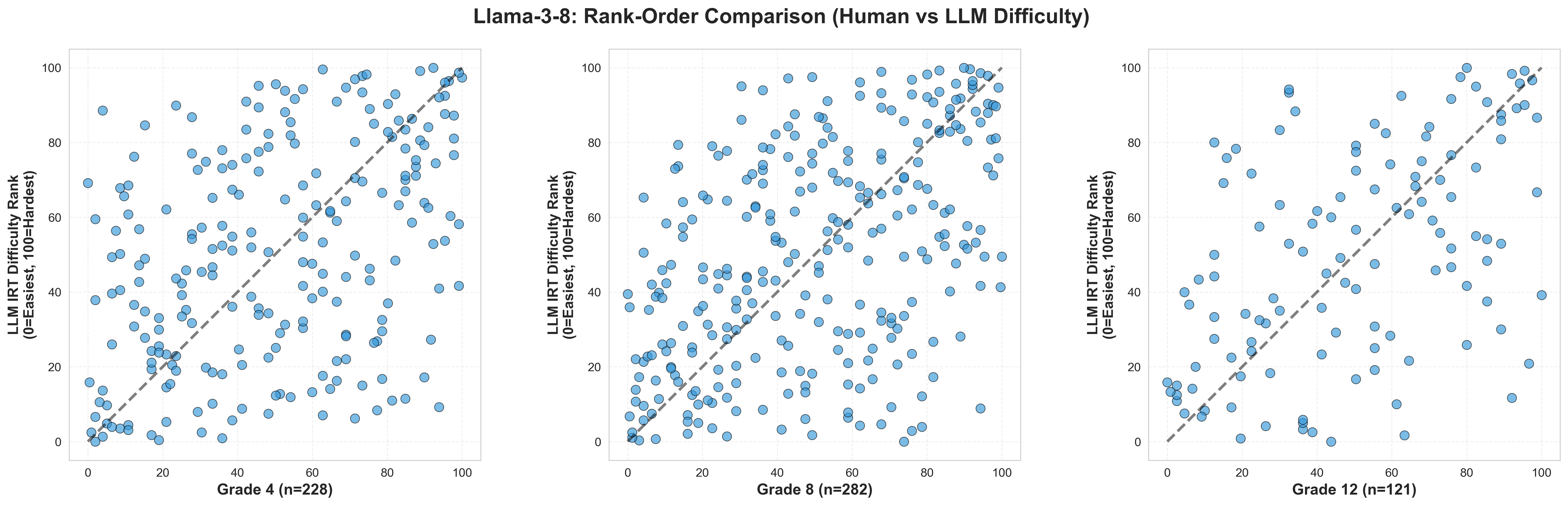}
    
    \caption{Complete rank-order difficulty scatter plots for select models \textit{( Gemma-2-9, Gemma-3-27, Llama-3-8)} across three grade levels. Each panel shows the rank-ordered difficulty of 631 test items (0=easiest, 100=hardest) for human students (x-axis) versus LLM simulations (y-axis).  Tighter clustering around the diagonal indicates better preservation of relative item difficulty. These visualizations confirm that correlation strength reliably predicts visual clustering quality, with r>0.65 indicating suitable performance for preliminary item screening.}
    \label{fig:scatter_plots}
\end{figure*}
\begin{table*}[h!]
\centering
\begin{tabular}{lccccccccc}
\toprule
\multirow{2}{*}{\textbf{Model}} & \multicolumn{3}{c}{\textbf{(r, PC) $\uparrow$}} & \multicolumn{3}{c}{\textbf{($\rho$, SC) $\uparrow$}} \\
\cmidrule(lr){2-4} \cmidrule(lr){5-7}
 & \textbf{G4} & \textbf{G8} & \textbf{G12} & \textbf{G4} & \textbf{G8} & \textbf{G12} \\
\midrule
GPT-4o & 0.15 & 0.06 & 0.01 & 0.14 & 0.06 & 0.01 \\
\midrule
Gemma-2-9b & 0.003 & 0.017 & -0.120 & 0.058 & -0.048 & -0.067 \\
Gemma-2-27b & -0.042 & -0.094 & -0.018 & -0.003 & -0.091 & -0.031 \\
\midrule
Gemma-3-4b & 0.059 & -0.079 & 0.045 & 0.053 & -0.097 & -0.035 \\
Gemma-3-12b & 0.020 & 0.014 & 0.108 & 0.006 & 0.010 & -0.105 \\
Gemma-3-27b & 0.021 & -0.074 & -0.059 & 0.021 & -0.087 & -0.053 \\
\midrule
Llama-3-8b & -0.011 & 0.031 & -0.066 & -0.037 & 0.027 & -0.048 \\
Llama-3.3-70b & 0.095 & 0.029 & -0.014 & 0.098 & 0.036 & -0.053 \\
\midrule
Qwen2.5-14B & 0.071 & 0.019 & -0.034 & 0.062 & 0.022 & -0.029 \\
Qwen2.5-32B & 0.100 & -0.042 & -0.050 & 0.137 & 0.002 & -0.068 \\
\midrule
Qwen3-4b & 0.010 & -0.058 & 0.041 & 0.022 & -0.047 & 0.024 \\
\bottomrule
\end{tabular}
\caption{Direct Prompting Correlation Results by Model and Grade. \textit{r} = Pearson correlation; $\rho$ = Spearman correlation; PC = Pearson Correlation; SC = Spearman Correlation. \textit{GPT-4o was evaluated under direct prompting only.}} 
\label{tab:direct_results}
\end{table*}

\begin{table*}[h!]
\centering
\begin{tabular}{lccccccccc}
\toprule
\multirow{2}{*}{\textbf{Model}} & \multicolumn{3}{c}{\textbf{(r, PC) $\uparrow$}} & \multicolumn{3}{c}{\textbf{($\rho$, SC) $\uparrow$}} \\
\cmidrule(lr){2-4} \cmidrule(lr){5-7}
 & \textbf{G4} & \textbf{G8} & \textbf{G12} & \textbf{G4} & \textbf{G8} & \textbf{G12} \\
\midrule
Gemma-2-9b & 0.028 & -0.002 & -0.139 & 0.041 & -0.017 & -0.108 \\
Gemma-2-27b & -0.018 & -0.080 & -0.015 & -0.016 & -0.063 & -0.035 \\
\midrule
Gemma-3-4b & 0.031 & -0.001 & -0.018 & 0.000 & -0.035 & -0.039 \\
Gemma-3-12b & 0.033 & 0.027 & -0.110 & 0.022 & 0.016 & -0.124 \\
Gemma-3-27b & 0.001 & -0.079 & -0.037 & -0.011 & -0.084 & -0.016 \\
\midrule
Llama-3-8b & -0.015 & 0.037 & -0.029 & -0.051 & 0.016 & -0.048 \\
Llama-3.3-70b & 0.087 & 0.021 & -0.038 & 0.087 & 0.046 & -0.039 \\
\midrule
Qwen2.5-14B & 0.077 & -0.021 & -0.048 & 0.066 & -0.027 & -0.055 \\
Qwen2.5-32B & 0.094 & -0.059 & 0.006 & 0.117 & -0.040 & 0.000 \\
\midrule
Qwen3-4b & -0.008 & -0.018 & 0.054 & 0.003 & -0.033 & 0.044 \\
\bottomrule
\end{tabular}
\caption{Averaged Prompting Correlation Results by Model and Grade. \textit{r} = Pearson correlation; $\rho$ = Spearman correlation; PC = Pearson Correlation; SC = Spearman Correlation}
\label{tab:averaged_results}
\end{table*}
\begin{table*}[h!]
\centering
\small
\begin{tabular}{llcccccccccc}
\toprule
\multirow{2}{*}{\textbf{Grade}} & \multirow{2}{*}{\textbf{Classification}} & \multicolumn{10}{c}{\textbf{Spearman Correlation ($\rho$) $\uparrow$}} \\
\cmidrule(lr){3-12}
 & & \textbf{G2-9} & \textbf{G2-27} & \textbf{G3-4} & \textbf{G3-12} & \textbf{G3-27} & \textbf{L3-8} & \textbf{L3-70} & \textbf{Q2.5-14} & \textbf{Q2.5-32} & \textbf{Q3-4} \\
\midrule
\multirow{5}{*}{\textbf{4}} 
 & Algebra & 0.74 & 0.78 & 0.81 & 0.80 & 0.78 & 0.54 & 0.67 & 0.66 & 0.74 & 0.67 \\
 & Data Analysis & 0.62 & 0.50 & 0.51 & 0.54 & 0.77 & 0.10 & 0.36 & 0.52 & 0.37 & 0.36 \\
 & Geometry & 0.74 & 0.83 & 0.73 & 0.71 & 0.59 & 0.40 & 0.59 & 0.67 & 0.40 & 0.58 \\
 & Measurement & 0.81 & 0.75 & 0.69 & 0.78 & 0.80 & 0.60 & 0.72 & 0.72 & 0.75 & 0.63 \\
 & Number Properties & 0.59 & 0.61 & 0.36 & 0.48 & 0.60 & 0.35 & 0.47 & 0.45 & 0.55 & 0.62 \\
\midrule
\multirow{5}{*}{\textbf{8}} 
 & Algebra & 0.45 & 0.60 & 0.35 & 0.61 & 0.64 & 0.30 & 0.49 & 0.47 & 0.56 & 0.43 \\
 & Data Analysis & 0.69 & 0.68 & 0.59 & 0.78 & 0.84 & 0.59 & 0.56 & 0.58 & 0.60 & 0.56 \\
 & Geometry & 0.45 & 0.57 & 0.45 & 0.63 & 0.64 & 0.34 & 0.50 & 0.54 & 0.49 & 0.56 \\
 & Measurement & 0.77 & 0.79 & 0.71 & 0.78 & 0.77 & 0.47 & 0.61 & 0.80 & 0.73 & 0.69 \\
 & Number Properties & 0.65 & 0.71 & 0.60 & 0.72 & 0.76 & 0.47 & 0.54 & 0.60 & 0.66 & 0.63 \\
\midrule
\multirow{5}{*}{\textbf{12}} 
 & Algebra & 0.64 & 0.71 & 0.69 & 0.70 & 0.70 & 0.53 & 0.37 & 0.47 & 0.35 & 0.58 \\
 & Data Analysis & 0.80 & 0.80 & 0.72 & 0.89 & 0.74 & 0.38 & 0.29 & 0.74 & 0.55 & 0.62 \\
 & Geometry & 0.57 & 0.61 & 0.42 & 0.51 & 0.69 & 0.05 & 0.43 & 0.21 & 0.55 & 0.13 \\
 & Measurement & 0.87 & 0.86 & 0.59 & 0.81 & 0.88 & 0.49 & 0.53 & 0.77 & 0.71 & 0.65 \\
 & Number Properties & 0.80 & 0.86 & 0.72 & 0.76 & 0.73 & 0.52 & 0.56 & 0.58 & 0.58 & 0.77 \\
\bottomrule
\end{tabular}
\caption{Spearman Correlations by Content Area and Grade Level Across Models. Model abbreviations: G2/G3 = Gemma-2/3, L3 = Llama-3, Q2.5/Q3 = Qwen2.5/3. Numbers indicate parameter count in billions (e.g., G2-9 = Gemma-2-9b). To understand where simulations succeed and fail, we analyze correlations across 
the five content areas.  From the table, we observe that problems that evaluate students understanding on the Measurement content area, consistently achieve relatively stronger correlations. Problems on Algebra and Geometry show the weakest performance.}
\label{tab:content_area_results}
\end{table*}

\begin{table*}[h!]
\centering
\small
\begin{tabular}{llcccccccccc}
\toprule
\multirow{2}{*}{\textbf{Grade}} & \multirow{2}{*}{\textbf{Classification}} & \multicolumn{10}{c}{\textbf{Distractor Match (\%) $\uparrow$}} \\
\cmidrule(lr){3-12}
 & & \textbf{G2-9} & \textbf{G2-27} & \textbf{G3-4} & \textbf{G3-12} & \textbf{G3-27} & \textbf{L3-8} & \textbf{L3-70} & \textbf{Q2.5-14} & \textbf{Q2.5-32} & \textbf{Q3-4} \\
\midrule
\multirow{5}{*}{\textbf{4}} 
 & Algebra & 43.3 & 43.3 & 26.7 & 33.3 & 43.3 & 46.7 & 36.7 & 50.0 & 36.7 & 40.0 \\
 & Data Analysis & 44.0 & 48.0 & 40.0 & 56.0 & 48.0 & 44.0 & 40.0 & 48.0 & 48.0 & 28.0 \\
 & Geometry & 36.8 & 36.8 & 31.6 & 31.6 & 42.1 & 31.6 & 31.6 & 36.8 & 21.1 & 52.6 \\
 & Measurement & 27.8 & 46.3 & 46.3 & 38.9 & 29.6 & 51.9 & 50.0 & 42.6 & 40.7 & 53.7 \\
 & Number Properties & 34.0 & 29.0 & 34.0 & 32.0 & 20.0 & 40.0 & 43.0 & 30.0 & 37.0 & 38.0 \\
\midrule
\multirow{5}{*}{\textbf{8}} 
 & Algebra & 27.8 & 29.2 & 26.4 & 29.2 & 45.8 & 22.2 & 30.6 & 27.8 & 33.3 & 33.3 \\
 & Data Analysis & 33.3 & 42.4 & 30.3 & 30.3 & 30.3 & 27.3 & 18.2 & 45.5 & 30.3 & 42.4 \\
 & Geometry & 28.9 & 31.1 & 40.0 & 35.6 & 28.9 & 31.1 & 35.6 & 33.3 & 26.7 & 35.6 \\
 & Measurement & 41.7 & 52.1 & 33.3 & 45.8 & 35.4 & 37.5 & 37.5 & 25.0 & 45.8 & 35.4 \\
 & Number Properties & 34.5 & 38.1 & 36.9 & 42.9 & 45.2 & 27.4 & 29.8 & 42.9 & 44.0 & 28.6 \\
\midrule
\multirow{5}{*}{\textbf{12}} 
 & Algebra & 31.4 & 25.7 & 40.0 & 42.9 & 48.6 & 25.7 & 17.1 & 25.7 & 25.7 & 34.3 \\
 & Data Analysis & 42.1 & 42.1 & 36.8 & 42.1 & 26.3 & 10.5 & 10.5 & 42.1 & 31.6 & 42.1 \\
 & Geometry & 50.0 & 50.0 & 42.9 & 35.7 & 21.4 & 35.7 & 50.0 & 14.3 & 35.7 & 35.7 \\
 & Measurement & 26.1 & 30.4 & 43.5 & 39.1 & 39.1 & 47.8 & 34.8 & 43.5 & 34.8 & 21.7 \\
 & Number Properties & 36.7 & 46.7 & 40.0 & 50.0 & 46.7 & 20.0 & 23.3 & 50.0 & 50.0 & 26.7 \\
\bottomrule
\end{tabular}
\caption{Distractor Match by Content Area and Grade Level Across Models. Model abbreviations: G2/G3 = Gemma-2/3, L3 = Llama-3, Q2.5/Q3 = Qwen-2.5/3. Numbers indicate parameter count in billions.}
\label{tab:distractor_match}
\end{table*}

\begin{figure*}
    \centering
    \includegraphics[width=0.8\linewidth]{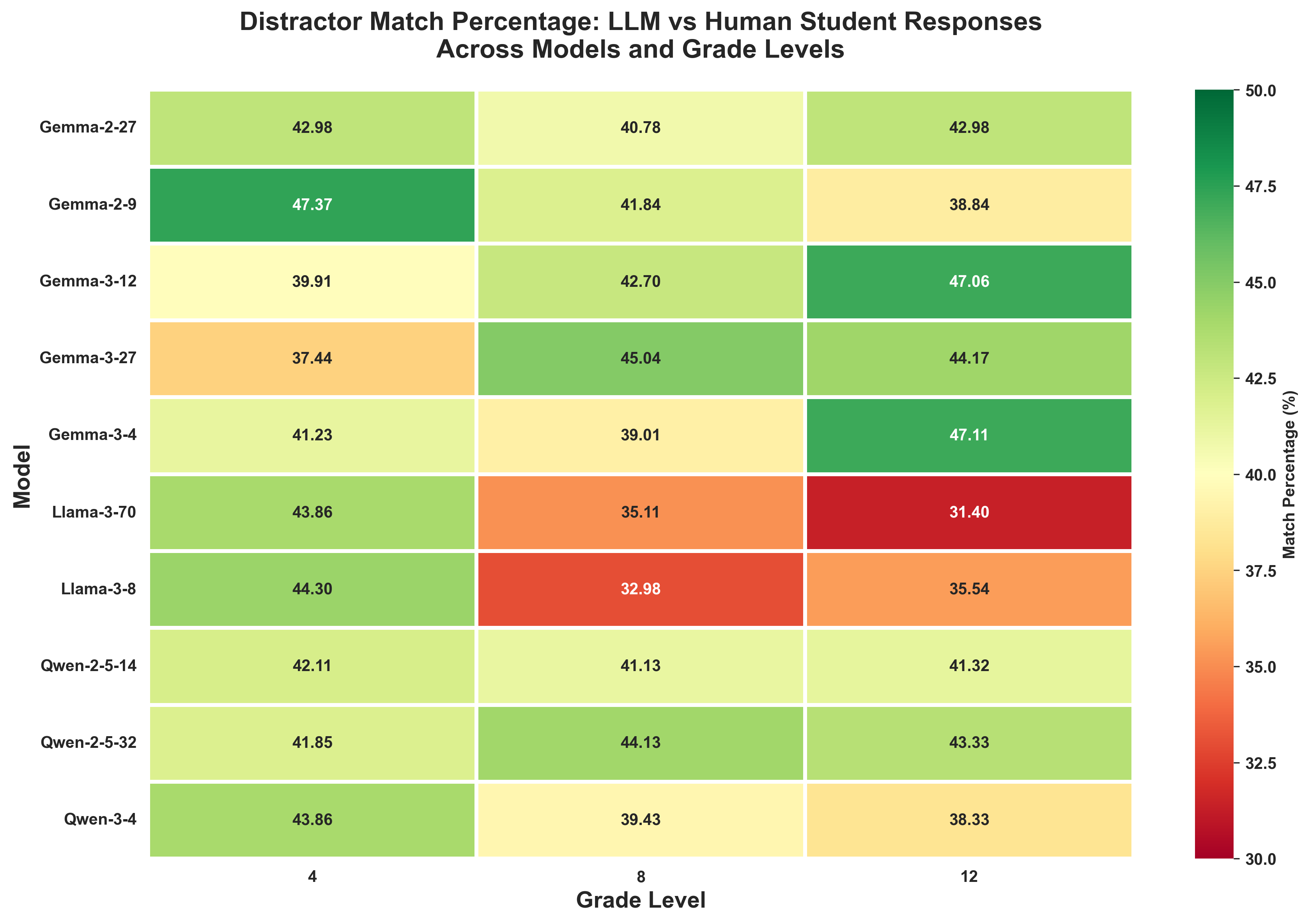}
    \caption{Full report of top distractors that matches for the real-world and llm response. The percentage shows the exact match on the top-1 distractor answers selected by our simulated students and the real world students.}
    \label{fig:distractors}
\end{figure*}
\begin{table*}
\centering

\label{tab:student_responses}
\begin{tabular}{|l|p{0.7\linewidth}|}
\hline
\textbf{Question} & \textbf{Kirstin wants to buy a flute that costs \$240. She has saved \$20 each week for 3 weeks. How many more weeks does Kirstin need to save money if she continues to save \$20 each week?}

 \\
\hline
\textbf{Answer Choices} &A. 9 weeks B. 10 weeks C. 11 weeks D. 12 weeks
 \\
\hline
\textbf{Content Area} & Number Properties and Operations \\
\hline
\textbf{Below Basic Response} & \{'\textbf{reasoning}': 'She saved 20 for 3 weeks which means 20 times 3, so that is 50 dollars. I need to figure out the difference between 240 and 50. I think that is 190. Hmm...If she saves 20 each week, that is 180 subtracted by 20, which is 160 weeks, I don't see the 160 so I will pick the highest option which is 12', 'answer key': 'D'\}\\  
\hline
\textbf{Basic Response} &  \{'\textbf{reasoning}': 'She saved \$60. I think I need to take \$240 and subtract \$60...that is \$180 left.  I need to figure out how many more weeks it will take to save \$180. \$20 each week...  so maybe 9 weeks?', 'answer key': 'A'\}\\
\hline
\textbf{Proficient Response} &  \{'\textbf{reasoning}': "Kirstin has saved 3 weeks * \$20/week = \$60. She needs \$240 - \$60 = \$180 more. If she saves \$20 each week, she'll need to save for \$180 / \$20/week = 9 more weeks.", 'answer key': 'A'\}\\
\hline
\textbf{Advanced Response} & \{'\textbf{reasoning}': 'First, I need to figure out how much money Kirstin has saved so far. She saves \$20 each week for 3 weeks, so she has \$20 x 3 = \$60.  Next, I need to figure out how much more money she needs. The flute costs \$240 and she has \$60 saved, so she needs \$240 - \$60 = \$180 more. Finally, to figure out how many more weeks she needs to save, I divide the amount she needs by how much she saves each week: \$180 / \$20 = 9 weeks.', 'answer key': 'A'\}\\
\hline
\end{tabular}
\caption{Sample student responses from Gemma-2-9B across skill levels for Grade 4 question item.}
\label{tab:sample_answer_1}
\end{table*}

\begin{table*}
\centering

\label{tab:student_responses}
\begin{tabular}{|l|p{0.7\linewidth}|}
\hline
\textbf{Question} & \textbf{Joseph's current test scores, in points, are 83, 97, 80, 87, and 93. He earns 100 points on his sixth test. By how many points does Joseph's mean (average) test score increase with the sixth test?}

 \\
\hline
\textbf{Answer Choices} &A. 2  B. 3  C. 6  D. 12  E. 20
 \\
\hline
\textbf{Content Area} & Data Analysis, Statistics, and Probability \\
\hline
\textbf{Below Basic Response} & \{'\textbf{reasoning}': '', 'answer key': ''\}\\  
\hline
\textbf{Basic Response} &  \{'\textbf{reasoning}': 'Let’s find out how many points Joseph had before his sixth test. … If I add them together, he has 440 points. The sixth test added 100 points, so now he has 540 points total… To find the average, I divide 540 by 6, which is 90. Before the sixth test, 440/5 = 88. He’s increased by 90 minus 88, so he increased by two points', 'answer key': 'A'\}\\
\hline
\textbf{Proficient Response} &  \{'\textbf{reasoning}': "First five scores sum to 430 points (so 430/5 = 86 as the old average). Adding the 100-point test gives 530 total. Dividing by 6 gives a new average of 88.33. Subtracting, the difference is 2.33, so the mean increases by 2 points (rounding to whole points).", 'answer key': 'A'\}\\
\hline
\textbf{Advanced Response} & \{'\textbf{reasoning}': 'Calculating the original total: 83+97+80+87+93 = 440. Dividing by 5 gives 88, New total with sixth score: 440+100 = 540. Dividing 540 by 6 yields 90. Subtracting yields 2, so the average increases by 2 points.', 'answer key': 'A'\}\\
\hline
\end{tabular}
\caption{Sample student responses from Gemma-2-9B across skill levels for Grade 8 question item.}
\label{tab:sample_answer_2}
\end{table*}

\begin{table*}
\centering

\label{tab:student_responses}
\begin{tabular}{|l|p{0.7\linewidth}|}
\hline
\textbf{Question} & \textbf{Which of the following expressions does NOT have the same value as two times five times eight?}

 \\
\hline
\textbf{Answer Choices} & A. two times forty; B. ten times sixteen; C. five times eight times two; D. two times five times two times four
 \\
\hline
\textbf{Content Area} & Number Properties and Operations \\
\hline
\textbf{Below Basic Response} &  \{'\textbf{reasoning1}': 'I think that is fifty minus twelve...', 'answer key': 'B', '\textbf{reasoning2}': 'I know two times five times eight equals forty. I think two times forty is the only choice that doesn't have forty in it.', 'answer key': 'B'\}\\
\hline
\textbf{Basic Response} &  \{'\textbf{reasoning1}': 'I know 2 times 5 is 10, and 5 times 8 is 40, so I thought 10 times 40. I see the answer is 200, so I thought I did something wrong, and I checked the other answers. I didnt see any that were really wrong, but I decided to pick the one that was different. It was the only one that had the word times twice, so I picked that one', 'answer key': 'D', '\textbf{reasoning2}': 'I know two times five times eight equals forty. I think two times forty is the only choice that doesn't have forty in it.', 'answer key': 'B'\}\\
\hline
\textbf{Proficient Response} &  \{'\textbf{reasoning1}': 'I know that 2 times 5 times 8 is 80, so I need to figure out which one is not 80. I know that 2 times 40 is 80, so that's not the answer. I know that 10 times 16 is 160, so that's not the answer. I know that 5 times 8 times 2 is 80, so that's not the answer. I know that 2 times 5 times 2 times 4 is 80, so that's not the answer. ', 'answer key': 'B', '\textbf{reasoning2}': 'I know that 2 times 5 times 8 is 80, so I need to figure out which one is not 80. I know that 2 times 40 is 80, so that's not the answer. I know that 10 times 16 is 160, so that's not the answer. I know that 5 times 8 times 2 is 80, so that's not the answer. I know that 2 times 5 times 2 times 4 is 80, so that's not the answer. I think the answer is A.', 'answer key': 'B'\}\\
\hline
\textbf{Advanced Response} & \{'\textbf{reasoning1}': 'The first step is to find two times five, which is ten. Then, I can multiply ten by eight to get eighty. Since two times five times eight is equal to eighty, I will eliminate any answer choices that equal eighty. The only answer choice that does not equal eighty is two times forty, so I will choose this answer. I also notice that the other answer choices can be simplified to have factors of five and eight, so this choice must be the right one. I could also double-check by plugging in numbers, but this would take longer.', 'answer key': 'A', '\textbf{reasoning2}': 'I can see that two times five times eight is 40, so I need to find the expression that doesn't also equal 40. I notice that ten times sixteen is 160. So it is not the same as the other three choices, which are 40, 80, and 80. So my answer is B.', 'answer key': 'B'\}\\
\hline
\end{tabular}
\caption{Sample student responses from Llama-3-8B across skill levels for Grade 4 question item. This model selects B despite flawed reasoning. Below Basic claims "2×5×8=40" (wrong =80); Proficient calculates correctly, states "answer is A," yet picks B. This model prioritizes accuracy over role-fidelity, generating superficial errors while making correct selections.}
\label{tab:sample_answer_3}
\end{table*}

\end{document}